%% file: main.tex
\documentclass[10pt,twocolumn,letterpaper]{article}

\usepackage[pagenumbers]{iccv} 

\usepackage{tikz}
\usetikzlibrary{shapes,snakes}

\usepackage{fancyhdr}
\usepackage{graphicx}
\usepackage{algorithm} 
\usepackage{algpseudocode} 
\usepackage{mathtools}
\usepackage{dsfont}
\usepackage{url}
\usepackage{amssymb}
\usepackage{xcolor} 

\usepackage{comment}
\usepackage{bm}
\usepackage{amsmath}
\usepackage{multicol}
\usepackage{float}
\usepackage{changepage}
\usepackage[most]{tcolorbox}
\usepackage{adjustbox}

\usepackage{pgffor}


\definecolor{iccvblue}{rgb}{0.21,0.49,0.74}
\usepackage[pagebackref,breaklinks,colorlinks,allcolors=iccvblue]{hyperref}

\def\paperID{12851} 
\def\confName{ICCV}
\def\confYear{2025}

\title{ Disentanglement in T-space for Faster and Distributed Training of Diffusion Models with Fewer Latent-states }

\author{Samarth Gupta \\ Amazon  \\ {\tt\small samargpt@amazon.com}
\and    Raghudeep Gadde  \\ Amazon  \\ {\tt\small rggadde@amazon.com} \and    Rui Chen \\ Amazon   \\ {\tt\small chenrui@amazon.com}  
\and  Aleix M. Martinez \\ Amazon  \\ {\tt \small maleix@amazon.com}
}

\begin{document}

\twocolumn[{%
\renewcommand\twocolumn[1][]{#1}%
\maketitle
\vspace{-0.5cm}
\includegraphics[width=\textwidth]{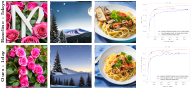}
\captionof{figure}{\label{fig:teaser} Samples from a text-to-image model trained from scratch for a day using the proposed disentanglement in T-space Vs baseline model with 1000 latent-states trained over 5 days. Prompts used are  \texttt{Close up of letter K made out of pink roses},  \texttt{Snow covered mountain with tall trees in the foreground and shining stars in the background} \& \texttt{Dish of pasta and seafood served in white plate on a blue tablecloth}. \textbf{Last Column}: Convergence plots with TIFA and CLIP scores. Proposed method provides  ~6x to 8x speedups over baseline.}
\vspace{1cm}
}]

%
%
%

\input{sec/0_abstract}    
\input{sec/1_intro}
\vspace{-0.12cm}
\input{sec/2_diffusion_models}

\input{sec/3_fewer_latent-states}
\input{sec/4_disentanglement}

{
    \small
    \bibliographystyle{ieeenat_fullname}
    \bibliography{main}
}

\clearpage
 \appendix
\input{sec/X_suppl}

\end{document}

%% file: sec/0_abstract.tex

\begin{abstract}
We challenge a fundamental assumption of diffusion models, namely, that a large number of latent-states or time-steps is required for training so that the reverse generative process is close to a Gaussian. 
We first show that with careful selection of a noise schedule, diffusion models trained over a small number of latent states (i.e. $T \sim 32$) match the performance of models trained over a much large number of latent states ($T \sim 1,000$). Second, we push this limit (on the minimum number of latent states required) to a single latent-state, which we refer to as complete disentanglement in T-space. We show that high quality samples can be
easily generated by the disentangled model obtained by combining several independently trained  single latent-state models.
We provide extensive experiments to show that the proposed disentangled model provides 4-6$\times$ faster convergence measured across a variety of metrics on two different datasets. 
\end{abstract}

%% file: sec/1_intro.tex
\section{Introduction}\label{sec:intro}
Diffusion models \cite{ho_ddpm, dm_beat_gan, stable_diff}  have made impressive progress in synthesizing near perfect photo-realistic images. These include Stable-Diffusion (SD) i.e. SD-2.1, SD-XL \cite{sdxl},  SD-3 \cite{sd3} amongst many others. 
In diffusion modeling, noise is gradually added to the data as a part of the forward process.
During training to learn the reverse-process, the denoising module (\eg, U-Net \cite{U-net} or DiT \cite{dit_peebles} ) is trained on a batch of noisy samples at varied SNR levels corresponding to different time-steps.
Importantly, a large number of time-steps or latent-states are used in practice such that the reverse process is close to Gaussian.
As this core assumption has gone mainly unchallenged, it has become common practice to train models with  large number of  latent states.

Interestingly, despite the diffusion model being trained over a large number of latent-states, but at inference, the  model is employed to de-noise only at a handful of latent-states $\sim 32$, thanks to faster sampling methods like DDIM \cite{ddim}. Motivated by this observation, we challenge the fundamental assumption. We show that with careful selection of noise schedule, diffusion models can be easily trained over much fewer number of latent-states without any loss in performance in terms of both convergence and the final sample quality. 
To the best of our knowledge, we are the first to achieve such feat.

We push the limit to reduce the number of latent-states all the way to 1! We refer to this as \emph{complete disentanglement in T-space}. We show that it is feasible to train models only on a single latent state and high-quality samples can be generated by combining these models at inference. Importantly, disentangled model has the following \textbf{advantages}:
\begin{enumerate*}[label=(\roman*)]
\item higher convergence due to linear scaling in throughput,
\item improved final performance,
\item increased model capacity,
\item almost negligible or marginal gain in sampling time at inference despite overall large model sizes,
\item distributed training provides flexibility to train across different geographical regions and hardware.
\end{enumerate*}

We provide extensive experiments on two datasets: \begin{enumerate*}[label=(\roman*)] \item ImageNet and \item
a dataset of 700M image-text pairs to train a text-to-image (T2I) model. \end{enumerate*} Our disentangled T2I model trained for a day can match the performance of a baseline model trained over 5 days.

 \begin{figure*}
    \centering
    \includegraphics[width=\linewidth]{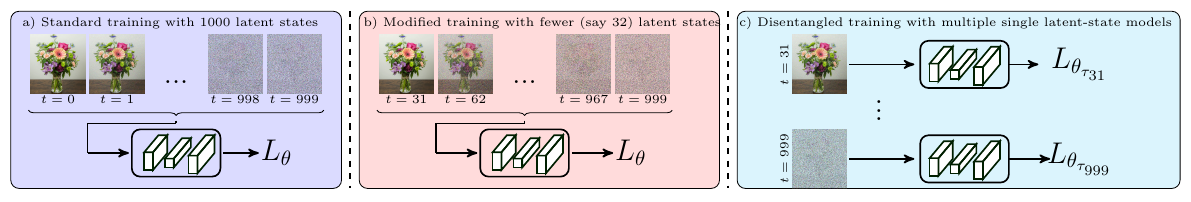} 
    \vspace{-0.8cm}
    \caption{\textbf{Overview}: We show that foundation diffusion models can be trained with fewer number of latent-states. The first block (a) shows the standard training procedure over 1000 latent states, i.e. algorithm \ref{algo:ddpm_training}. Next, (b) shows using algorithm \ref{algo:rev_training} we can train a diffusion model from scratch over much fewer latent states (for e.g. 32). Finally, in (c) using algorithm \ref{algo:limit_training}  we push this to an extreme where we train single-latent state models i.e. complete disentanglement over latent-states (or t-space) and combine these models to do inference.}
    \label{fig:overview}
    \vspace{-0.5cm}
\end{figure*}

%% file: sec/2_diffusion_models.tex
\section{Diffusion Models}\label{sec:diffusion_models}
\vspace{-0.1cm}
Denoising diffusion probabilistic models (DDPMs) \cite{dm_thermo,ho_ddpm} are a type of score-based generative models typically used to learn the underlying, unknown data-distribution $q(x_0)$ from a given set of data points (\ie, the training data, \eg, real-world images).
A forward process is defined such that noise is progressively added to an image $x_0$ for $T$ steps resulting in noisy images $x_1, \dots, x_T$. In particular, for a decreasing sequence $\alpha_{1:T} \in (0,1]^T$ with Gaussian transitions and sufficiently large $T$
the resulting process $q(x_t|x_{t-1})$ is a Markov chain. 
{\setlength{\abovedisplayskip}{2.5pt}
\setlength{\belowdisplayskip}{2.5pt}
\begin{equation}\label{eq:forw_process}
q(x_t | x_{t-1}) \coloneqq \mathcal{N}\left(x_t; \sqrt{\frac{\alpha_t}{\alpha_{t-1}}} x_{t-1}, \bigg(1 - \frac{\alpha_t}{\alpha_{t-1}}\bigg)I    \right)
\end{equation}
}
From \eqref{eq:forw_process}, it immediately follows that
{\setlength{\abovedisplayskip}{2.4pt}
\setlength{\belowdisplayskip}{2.4pt}   
\begin{equation}\label{eq:x_t_given_x_0}
  q(x_t | x_{0}) = \mathcal{N}\left(x_t; \sqrt{\alpha_t} x_{0}, \big(1 - \alpha_t)I \right),  
\end{equation}
}
i.e. we can generate a noisy image $x_t$ directly from $x_0$:
{\setlength{\abovedisplayskip}{2pt}
\setlength{\belowdisplayskip}{2pt}   
\begin{equation}
x_t = \sqrt{\alpha_t}x_0 + \sqrt{1-\alpha_t} \epsilon, \text{ where } \epsilon \sim \mathcal{N}(0,I).    
\end{equation}
}
When $\alpha_T$ is close to $0$, $q(x_T|x_0) $ tends to a standard Gaussian for all $x_0$.
A latent variable model $p_{\theta}(x_{0:T})$, with $x_1, \dots, x_T$ as latent states and  parameters $\theta$, is defined to  approximate the intractable reverse process $q(x_{t-1}|x_t)$. The conditionals $p_\theta(x_{t-1}|x_t)$ are modelled as Gaussians with fixed variances and mean functions with shared parameters $\theta$,
{\setlength{\abovedisplayskip}{2.5pt}
\setlength{\belowdisplayskip}{2.5pt}
\begin{subequations}\label{eq:reverse_process}
\begin{alignat}{10}
   & p_\theta(x_{t-1}|x_t) &&\coloneqq \mathcal{N}(x_{t-1}; \mu_{\theta}(x_{t-1},t), \sigma_{t}^2 I)   \\
   &p_\theta(x_T) && \coloneqq \mathcal{N}(0,I) .
\end{alignat}
\end{subequations}
}
It is important to discuss the settings under which the above definition of $p_\theta(x_{t-1}|x_t)$ can be a good approximation of the underlying true distribution for the reverse process $q(x_{t-1}|x_t)$. Using the Bayes rule, we can write
{\setlength{\abovedisplayskip}{2pt}
\setlength{\belowdisplayskip}{2pt} 
\begin{equation}\label{eq:bayes_q}
 q(x_{t-1}|x_t) \propto q(x_t|x_{t-1}) q(x_{t-1}) ,  
\end{equation}
}
where $q(x_{t-1})$ denotes the marginal distribution for state $x_{t-1}$. Feller in \cite{feller} showed that in the limit with infinitesimal step size, the product in equation \eqref{eq:bayes_q} primarily depends on $q(x_t|x_{t-1})$. Since the forward process $q(x_t|x_{t-1})$ is Gaussian by definition \eqref{eq:forw_process}, the reverse process becomes Gaussian too. Therefore, to ensure that the modelling assumption in equation  \eqref{eq:reverse_process} holds valid,  diffusion models typically use thousands of time-steps ($T \sim 10^3$), such that the individual step-sizes remain small.


\begin{algorithm}[t]
	\caption{DDPM Training \cite{ho_ddpm} } \label{algo:ddpm_training}
	\begin{algorithmic}[1]
                \Repeat
                \State ${x}_0 \sim q({x}_0),\; t  \sim \text{Uniform}(\{ 1, \dots, T \})$
                \State ${\epsilon} \sim \mathcal{N}({0}, {I}  )$ 
                \State $ L_{\theta} = \Vert  {\epsilon} - {\epsilon}_{\theta}( \sqrt{{\alpha}_t}{x}_0 + \sqrt{ 1 - {\alpha}_t }{\epsilon}, t)   \Vert^{2} $
                \State $ \theta \leftarrow \theta - \eta \nabla_{\theta} L_{\theta}$
                \Until{converged}
	\end{algorithmic} 
\end{algorithm}

We now focus on how the shared parameters $\theta$ are learned from data. The joint distribution $p_\theta(x_{0:T})$  from \eqref{eq:reverse_process} can be written as 
{\setlength{\abovedisplayskip}{-4pt}
\setlength{\belowdisplayskip}{1pt} 
\begin{equation}\label{eq:p_theta}
    p_\theta(x_{0:T}) = p_\theta(x_T)\prod_{t=1}^{T}p_\theta(x_{t-1}|x_t).
\end{equation}
}
In \eqref{eq:p_theta}, $p_{\theta}(x_{0:T})$ is a Markovian generative process and therefore can be used to synthesize samples $x_0$ by iteratively denoising $x_T$, \ie, pure Gaussian noise since  $p_\theta(x_T)$ is a standard Gaussian.  Finally, to learn the parameters $\theta$, a variational lower bound can be obtained on the log-likelihood, and thanks to the Gaussian modelling choice of the conditionals,
the variational lower bound reduces to the following simple loss function,
{\setlength{\abovedisplayskip}{3.5pt}
\setlength{\belowdisplayskip}{3.5pt}
\begin{equation}\label{eq:exp_loss}
    \mathds{E}_{x_0 \sim q(x_0),
    \epsilon \sim \mathcal{N}(0,I) }\Vert 
    {\epsilon}
    - {\epsilon}_{\theta}( \sqrt{{\alpha}_t}{x}_0 + \sqrt{ 1 - {\alpha}_t }{\epsilon}, t) 
    \Vert^{2} ,
\end{equation}
}
where $t$ is sampled uniformly from $\{1,\dots,T\}$ and ${\epsilon}_\theta(x_t,t)$ denotes the mean function\footnote{Note that predicting ${x}_0$ or ${\epsilon}$ are equivalent.} of the condition Gaussian $p_\theta(x_{t-1}|x_t)$.  The overall DDPM training procedure \cite{ho_ddpm}  is outlined in algorithm \ref{algo:ddpm_training}.
The reverse process is governed by equation \eqref{eq:ddpm_infer}.
{\setlength{\abovedisplayskip}{3pt}
\setlength{\belowdisplayskip}{3pt}
 \begin{alignat}{10}
               \hspace{0.5cm} {x}_{t-1} =& \sqrt{\frac{\alpha_{t-1}}{\alpha_t}} \bigg({x}_t - \frac{\alpha_{t-1} -\alpha_{t} }{\alpha_{t-1} \sqrt{1-\alpha_t}} {\epsilon}_{\theta}({x}_t,t) \bigg) + \notag \\
                &\sqrt{\frac{1-\alpha_{t-1}}{1-\alpha_t}\bigg( 1- \frac{\alpha_t}{\alpha_{t-1}} \bigg)      } \; \cdot\,{\epsilon}  \hspace{0.3cm}     \label{eq:ddpm_infer}             
                \end{alignat}
}
The detailed algorithm is provided in appendix. A large value of $T$ is chosen, typically $T \sim 10^3$, such that the modeling assumption of $p_\theta(x_{t-1}|x_t)$ as Gaussian holds valid. 
Thus, generating high-quality samples from pure-noise $x_T$
is slow due to the Markovian nature of equation \eqref{eq:ddpm_infer}.
\subsection{Non-Markovian Sampling: DDIM}
To overcome the slow sampling procedure of DDPMs, authors in \cite{ddim} extended the Markovian reverse sampling process of DDPMs to a more generalised non-Markovian sampling framework. They made a key observation that for a given $x_0$, the DDPM training objective can be expressed entirely in terms of the marginals, \ie, $q(x_{t-1}|x_0)$. Therefore, does not involve the joint distribution of the latent variables, \ie, $q(x_{1:T}|x_0)$. This combined with the fact that many different joint distributions have the same marginal distribution,
leads to a non-Markovian sampling process, while keeping the \textit{same training objective as} DDPMs. The resulting generalised sampling procedure \cite{ddim} is governed by equation \eqref{eq:ddim_T}.
{\setlength{\abovedisplayskip}{3pt}
\setlength{\belowdisplayskip}{3pt}
\begin{alignat}{10}
                \hspace{0.225cm} {x}_{t-1} =\;  & \sqrt{\alpha_{t-1}}\bigg( \frac{{x}_t - \sqrt{1-\alpha_t} {\epsilon}_{\theta}({x}_t,t)}{\sqrt{\alpha_t}} \bigg) +  \notag  \\
                & \sqrt{ 1-\alpha_{t-1} -\sigma_t^2}\cdot {\epsilon}_{\theta}({x}_t,t) + \sigma_t {\epsilon}  \label{eq:ddim_T}
                \end{alignat}
}
For a model trained with the DDPM objective of algorithm \ref{algo:ddpm_training}, different sampling schemes can be devised by carefully choosing the value of $\sigma_t$ in \eqref{eq:ddim_T}. 
In a deterministic setting where  
 $\sigma_t =0 \;\; \forall \;\; t \in \{1,\dots,T\}$,
the reverse process \eqref{eq:ddim_T}  is an implicit probabilistic model \cite{implicit_prob_model} and thus referred to as \textbf{denoising diffusion implicit model} (DDIM) \cite{ddim}.

Let $< \tau_1, \dots, \tau_\text{S}>$ be an increasing sub-sequence of $[1,\dots,T]$ of length $\text{S}$. For a forward process defined only on a subset of latent states $x_{\tau_1},\dots, x_{\tau_\text{S}}$, where the forward process still follows \eqref{eq:x_t_given_x_0}, \ie,
{\setlength{\abovedisplayskip}{3pt}
\setlength{\belowdisplayskip}{3pt} 
\begin{equation}
 q(x_{\tau_i}|x_0) = \mathcal{N}(x_{\tau_i}; \sqrt{\alpha_{\tau_i}}x_0, (1-\alpha_{\tau_i})I)  ,  \notag
\end{equation}}
the corresponding \textbf{accelerated} reverse process is governed by equation \eqref{eq:ddim_T_acc}
\begin{equation}
\footnotesize
{x}_{\tau_{i-1}} =  \bigg( \frac{{x}_{\tau_i} - \sqrt{1-\alpha_{\tau_i}} {\epsilon}_{\theta}({x}_{\tau_i},{\tau_i})}{\sqrt{\alpha_{\tau_i}/\alpha_{\tau_{i-1}}}   } \bigg) + \frac{ {\epsilon}_{\theta}({x}_{\tau_i},{\tau_i}) }{\sqrt{ (1-\alpha_{\tau_{i-1}})^{-1} }}
 \label{eq:ddim_T_acc}
\end{equation}

Importantly, \eqref{eq:ddim_T_acc} holds valid even for a model trained with the DDPM objective in \eqref{eq:exp_loss}.
Therefore, for a model trained with algorithm \ref{algo:ddpm_training}, 
samples can now be generated using \eqref{eq:ddim_T_acc}. This enables computing the reverse process only on a \textbf{small} subset of latent states of size $\text{S}$ instead of all $T$. 
Motivated by this we next work towards training diffusion models with fewer latent states.

%% file: sec/3_fewer_latent-states.tex
\section{Diffusion Models with Fewer Latent-States}\label{sec:motivation}
The main focus and one of the key contributions of this paper is to enhance our broader understanding of diffusion models by answering the following question:
\textbf{Can we effectively train diffusion models with fewer latent-states without compromising sample quality?}

Most common and widely used practice for training and inference of diffusion models comprises of the following:
\begin{itemize}
\item A model is trained with a large value of $T$, typically $T \sim 1,000$, \ie, with large number of latent variables   $x_1,\dots,x_T$ corresponding to $[1,\dots,T]$. 
\item At inference,  sampling is done only on a small subset of latent variables $x_{\tau_1},\dots, x_{\tau_\text{S}}$ of size $\text{S}$ using DDIM sampling i.e. \eqref{eq:ddim_T_acc}. In principle, $\tau_1, \dots, \tau_\text{S}$ may be any increasing sub-sequence of $[1,\dots, T]$. In practice, an arithmetic progression is commonly used with $\tau_{i}-\tau_{i-1} \approx  T/\text{S}$. 
For (foundational) diffusion models trained to learn an underlying  distribution, specifically of an image dataset, we observe that:
\begin{enumerate}
    \item \label{item:sampling_1}  The visual quality of generated sample deteriorates for very small arithmetic sub-sequences, \ie, to synthesize good quality samples $\text{S} \ge \text{S}_l$. In practice, $\text{S}_l \sim 32$, therefore an empirical lower-bound $\text{S}_l$ exists on $\text{S}$.
\item \label{item:sampling_2} There is no significant improvement in visual sample quality for very large arithmetic sub-sequences, \ie, $\text{S} \le \text{S}_u $ suffices. Typically $\text{S}_u \sim 64$, thus an empirical upper-bound $\text{S}_u$ also exists on $\text{S}$.
\item \label{item:sampling_3} The most commonly used setup is $\text{S} = 50$ with DDIM \ie when $\sigma_t =0$.    
\end{enumerate}
\end{itemize}
From observations \ref{item:sampling_1}-\ref{item:sampling_3} above, it is evident that even though \eqref{eq:ddim_T_acc} provides us with complete flexibility in choice of $\text{S}$ when sampling from a model trained with much larger  $T$, \ie,  $T \gg \text{S}$. However in practice $\text{S}$ is determined by (visual) sample quality. Consequently, $\text{S}$ typically remains fixed i.e. close to $\text{S}_u$ or at-most chosen from a small interval $[\text{S}_l,\text{S}_u]$.
Therefore high flexibility in choice of $\text{S}$ from \eqref{eq:ddim_T_acc} does not provide with any practical benefit, particularly when $\text{S} > \text{S}_u$. 

To answer the broader question that whether text-to-image diffusion model can be trained effectively with much fewer latent-states, we 
raise the following related question:

\tikzstyle{mybox} = [draw=red, fill=red!10, thick,
    rectangle, rounded corners, inner sep=8pt, inner ysep=8pt]

\begin{center}
\begin{tikzpicture}
\node [mybox] (box){%
    \begin{minipage}{0.92\linewidth}
\textbf{
\emph{Given that at inference the model denoises only for a handful of time-steps or SNR levels ($\sim32$), therefore can a model trained for only those corresponding time-steps  match the performance of a model trained over a much larger number of time-steps ($ \sim 10^3$) ?}
}
    \end{minipage}
};
\end{tikzpicture}%
\end{center}

An \textbf{affirmative} answer to the above question is our first major contribution in this paper. Subsequently in section \ref{sec:dis_t_space}  this will allow us to build a method to train models faster while achieving higher inference quality. 

The above question entails that the reverse process $q(x_{t-1}|x_t)$ is non-Gaussian and plausibly highly multi-modal. Given that $q(x_{t-1}|x_t)$ is non-Gaussian, the Gaussian assumption in \eqref{eq:reverse_process} need not hold. Crucially, our key observation regarding \eqref{eq:ddim_T_acc}  is that the only requirement for DDIM sampling to generate high-quality samples is that the model effectively denoises the input at various SNR levels ($\tau_1, \dots, \tau_\text{S}$), \textbf{irrespective of the total number of latent-states over which the model has been trained}.

This crucial observation allows us to propose algorithm \ref{algo:rev_training}. The most important part of algorithm \ref{algo:rev_training} is that the $\alpha_t$'s, which decide the noise-schedule in equations \eqref{eq:forw_process} and  \eqref{eq:x_t_given_x_0}, are computed assuming a large number of time-steps (\eg, $T=1,000$). As a result,   $\alpha_{\tau_1}, \dots, \alpha_{\tau_\text{S}}$ corresponding to $\tau_1, \dots, \tau_\text{S}$ in algorithm \ref{algo:rev_training} match their counterparts in algorithm \ref{algo:ddpm_training}. 
Matching  of $\alpha_t$'s between algorithm \ref{algo:ddpm_training} and  \ref{algo:rev_training} ensures that at inference, apart from  the model output $\epsilon_{\theta}(x_t,t)$, everything else remains the same in equation \eqref{eq:ddim_T}. 

\begin{algorithm}
	\caption{Fewer Latent-States DDPM Training (Ours) } \label{algo:rev_training}
	\begin{algorithmic}[1]
                 \State \textbf{Input:} $[\tau_1,\dots,\tau_\text{S}], T$
                \For{ $t \leftarrow [ 1,\dots,T ]$}
                    \State Compute $\alpha_t$
                \EndFor
                \Repeat
                \State ${x}_0 \sim q({x}_0), \; t  \sim \text{Uniform}( \{ \tau_1,\dots,\tau_\text{S} \})  $  
                \State ${\epsilon} \sim \mathcal{N}({0}, {I}  )$ 
                \State $ L_{\theta} = \Vert  {\epsilon} - {\epsilon}_{\theta}( \sqrt{{\alpha}_t}{x}_0 + \sqrt{ 1 - {\alpha}_t }{\epsilon}, t)   \Vert^{2} $
                \State $ \theta \leftarrow \theta - \eta \nabla_{\theta} L_{\theta}$
                \Until{converged}
                 \State \textbf{Output:} $\theta$

	\end{algorithmic} 
\end{algorithm}

\subsection{Experiments-I: Fewer latent-states}\label{sec:exp1}

 We now provide experiments to show that a diffusion model trained over fewer latent-state i.e. with a small number of DDPM time-steps using algorithm \ref{algo:rev_training}, matches the performance of a diffusion model trained with algorithm \ref{algo:ddpm_training}.
 We evaluate on two different setups (or datasets). The first setup is based on the standard ImageNet-1k  dataset comprised of 1.3M images. In the second case, we evaluate in a text-to-image foundation diffusion model setting on a dataset of 700M image-text pairs.
\subsubsection*{ImageNet} 
This dataset \cite{imagenet} comprises of 1.3 million labelled images belonging to 1000 different classes. The images are resized to $256\times256\times3$ resolution and then training is done in  the latent-space by computing embeddings of size $4\times32\times32$ using a vector-quantized (VQ) auto-encoder \cite{stable_diff, taming_transformers}. For all our ImageNet experiments, the diffusion model backbone is a class-conditioned U-Net with a total of $395$M trainable parameters \cite{stable_diff}.
 
We first train a baseline model using algorithm \ref{algo:ddpm_training}, with $T=1,000$ and a batch-size of $256$. We will compare all our models against this baseline. We next train four different models using algorithm \ref{algo:rev_training}, where each model is trained on $\text{S} = 8,16,32 \text{ and } 64$ latent-states, respectively. Apart from the number of latent-states each model has the same setup as the baseline in terms of model-architecture, dataset, batch-size, learning-rate etc. Its important to emphasize that the $\alpha_t$'s used for these models match with their counterpart of $1,000$ latent-states (\ie, the baseline).

To \textbf{evaluate} the performance of any model trained on ImageNet dataset we use Fréchet Inception Distance (FID) \cite{fid} and Inception Score (IS) \cite{improved_technique_gan}. To compute both of these  metrics  we generate $50$k synthetic images and 
compare them against the real-world images \cite{dm_beat_gan}. The evolution of  FID and IS during training for all the four fewer latent-state models along with the baseline is shown in figure \ref{fig:diff_latent_cin}. We provide generated samples for both the model types in Appendix.


\subsubsection*{Text-to-image (T2I) }
In this setup, we use a curated dataset of 700M high-quality image-text pairs \cite{scalability_agi}  which were supplemented with synthetic captions using BLIP2 \cite{blip2}. Similar to previous ImageNet experiments, all models discussed for this dataset are also trained from scratch to avoid any ambiguity unless mentioned otherwise.
\input{sec/fig_fewer}

For all our experiments, we use the U-Net from  Stable Diffusion (SD-2.1) with 865M parameters along with the OpenCLIP-H text encoder \cite{openclip, clip} with a token length of 77. To generate the input latents (\ie, image embedding), we use the pre-trained variational auto-encoder (VAE) of SD-2.1. 
To train all our models, we use a batch-size of $1,024$ at $256 \times 256 \times 3 $ resolution resulting in embeddings of size $ 4 \times 32 \times 32$.

Here again, we first train a baseline model (865M)  using algorithm \ref{algo:ddpm_training}, with $T=1,000$ and the same noise schedule as used in the publicly available codebase of SD-2.1. 
This model will be used as the baseline for comparison against all our subsequent models.
Using algorithm \ref{algo:rev_training}  with $\text{S} = 8,16,32 \text{ and } 64$ latent-states  we also train four different fewer latent-state models. Apart from the number of latent-states each model has the same setup as the baseline. 
We re-emphasize that the $\alpha_t$'s used for these fewer latent-state models match with baseline trained over $1,000$ latent-states.

We \textbf{evaluate} all our text-to-image models using various different metrics mainly CLIP \cite{openclip, clip} score to measure prompt alignment, CMMD \cite{cmmd} (a corrected version of FID) to measure the overall image fidelity. Note that FID has many drawbacks such as being biased, incorrect normality assumption and lack of diversity in Inception features, therefore is not well suited to evaluate T2I models.
CLIP and CMMD is calculated over a sample size of 30k using prompts from MSCOCO validation set \cite{mscoco}.
Finally, we also use TIFA \cite{tifa} scores which is based on visual question answering (VQA) methods and is calculated using a set of 4081 prompts provided by the author \cite{tifa}.
The evolution of CLIP, CMMD and TIFA scores during training for all these four models along with baseline is shown in figure \ref{fig:diff_latent_t2i}.


From figure \ref{fig:diff_latent}  we can conclude that \textbf{diffusion models trained on fewer latent-states (or DDPM steps) match both the convergence and the final sample quality of the baseline model trained on $1,000$ latent-states.} To the best of our knowledge, we are the first to achieve such results. Thus, conclusively  demonstrating that diffusion models can be trained on much fewer latent-states than previously thought. 

%% file: sec/fig_fewer.tex
\begin{figure*}[htbp]
    \centering
    \begin{subfigure}{\textwidth}
        \centering
        \begin{minipage}[b]{0.24\textwidth}
            \includegraphics[width=\textwidth]{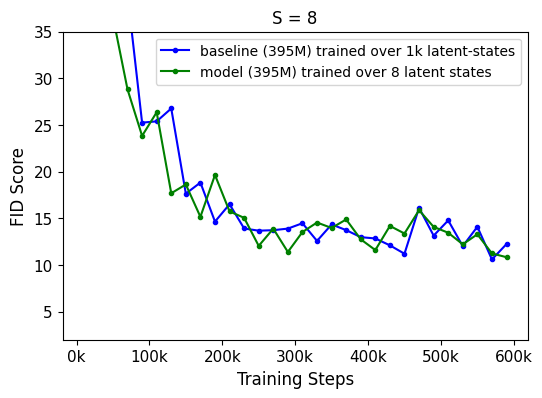}
        \end{minipage}
        \begin{minipage}[b]{0.24\textwidth}
            \includegraphics[width=\textwidth]{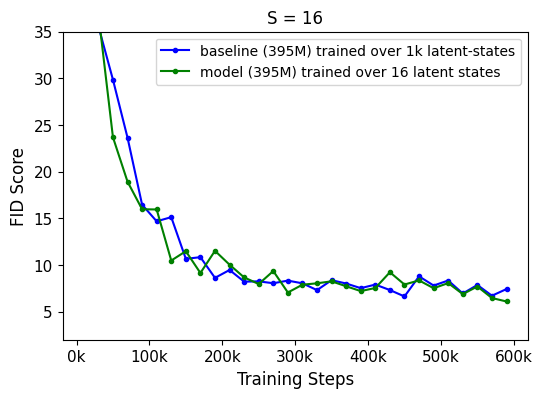}
        \end{minipage}
        \begin{minipage}[b]{0.24\textwidth}
            \includegraphics[width=\textwidth]{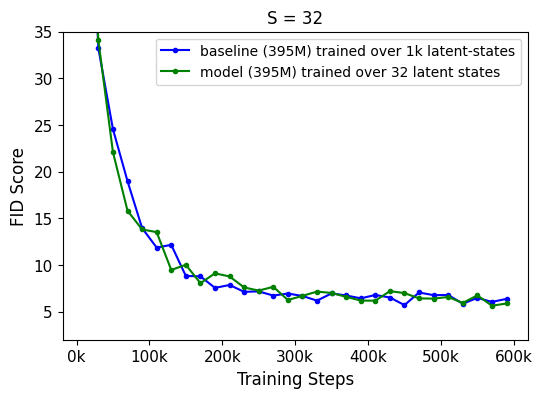}
        \end{minipage}
        \begin{minipage}[b]{0.24\textwidth}
            \includegraphics[width=\textwidth]{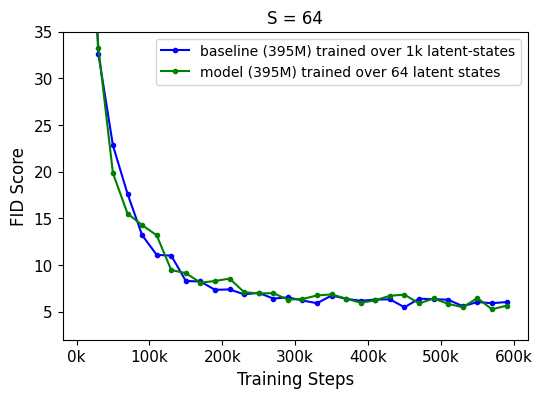}
        \end{minipage}
    \end{subfigure}    
    \vspace{4mm}
    \begin{subfigure}{\textwidth}
        \centering
        \begin{minipage}[b]{0.24\textwidth}
            \includegraphics[width=\textwidth]{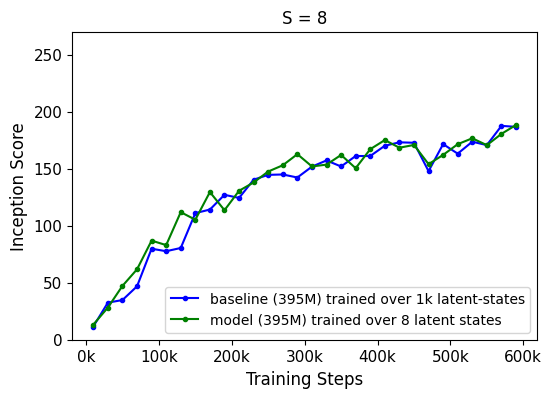}
        \end{minipage}
        \begin{minipage}[b]{0.24\textwidth}
            \includegraphics[width=\textwidth]{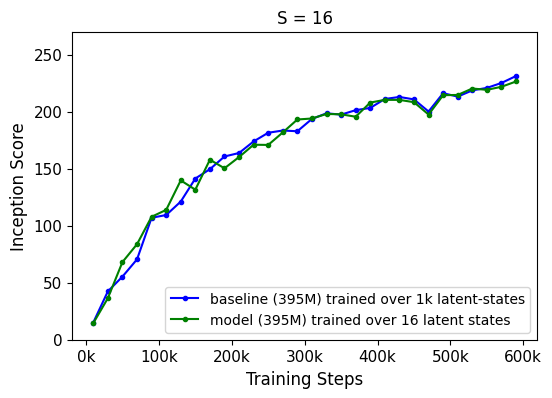}
        \end{minipage}
        \begin{minipage}[b]{0.24\textwidth}
            \includegraphics[width=\textwidth]{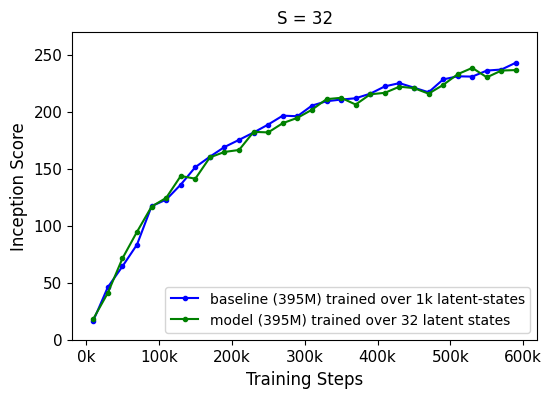}
        \end{minipage}
        \begin{minipage}[b]{0.24\textwidth}
            \includegraphics[width=\textwidth]{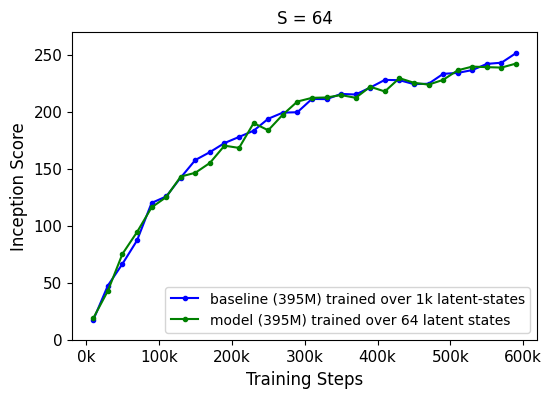}
        \end{minipage}
        \caption{ImageNet experiments with performance measured using FID and Inception scores.}\label{fig:diff_latent_cin}
        \vspace{-0.2cm}
    \end{subfigure}    
    \begin{subfigure}{\textwidth}
        \centering
        \begin{minipage}[b]{0.24\textwidth}
            \includegraphics[width=\textwidth]{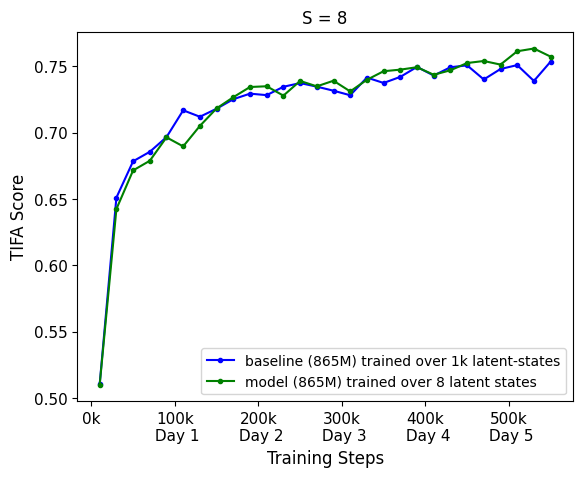}
        \end{minipage}
        \begin{minipage}[b]{0.24\textwidth}
            \includegraphics[width=\textwidth]{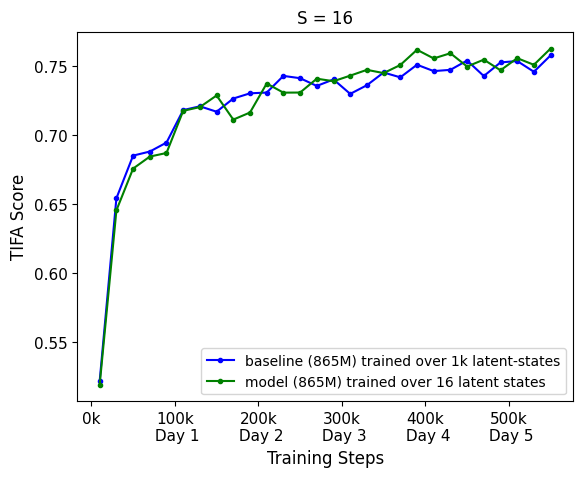}
        \end{minipage}
        \begin{minipage}[b]{0.24\textwidth}
            \includegraphics[width=\textwidth]{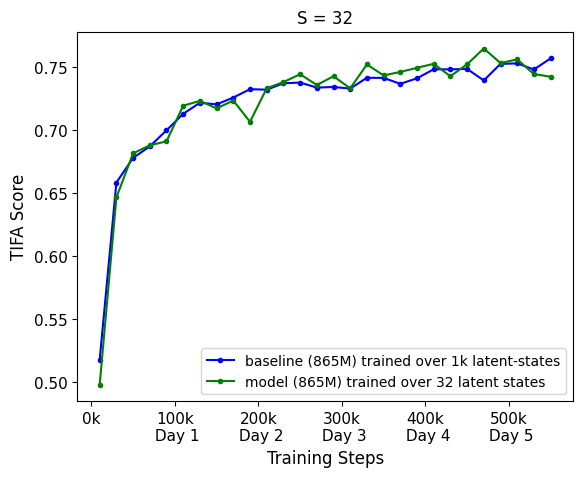}
        \end{minipage}
        \begin{minipage}[b]{0.24\textwidth}
            \includegraphics[width=\textwidth]{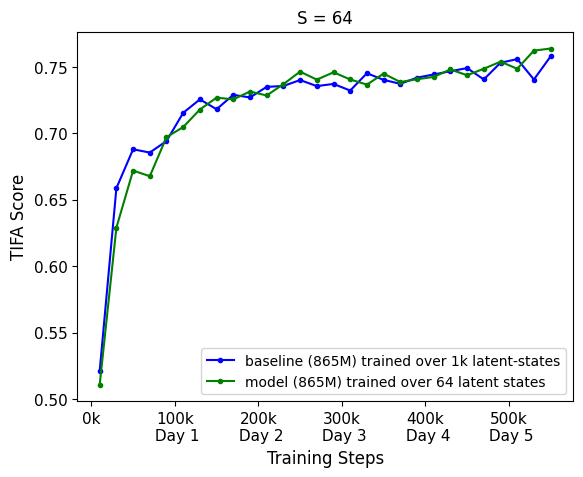}
        \end{minipage}
    \end{subfigure}
    \vspace{4mm}
    \begin{subfigure}{\textwidth}
        \centering
        \begin{minipage}[b]{0.24\textwidth}
            \includegraphics[width=\textwidth]{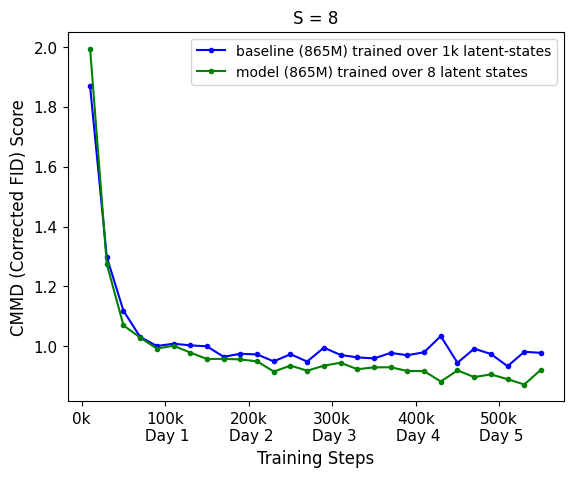}
        \end{minipage}
        \begin{minipage}[b]{0.24\textwidth}
            \includegraphics[width=\textwidth]{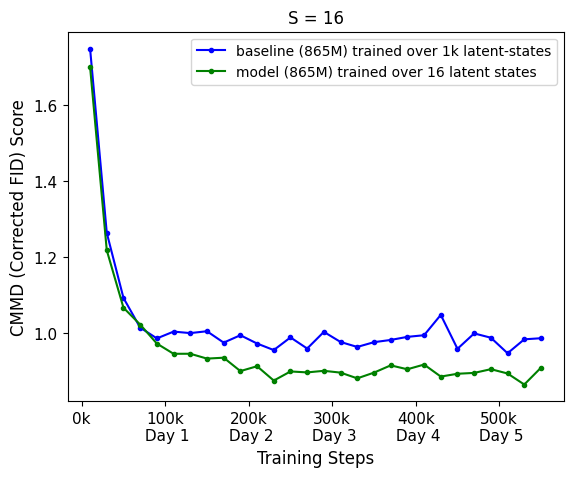}
        \end{minipage}
        \begin{minipage}[b]{0.24\textwidth}
            \includegraphics[width=\textwidth]{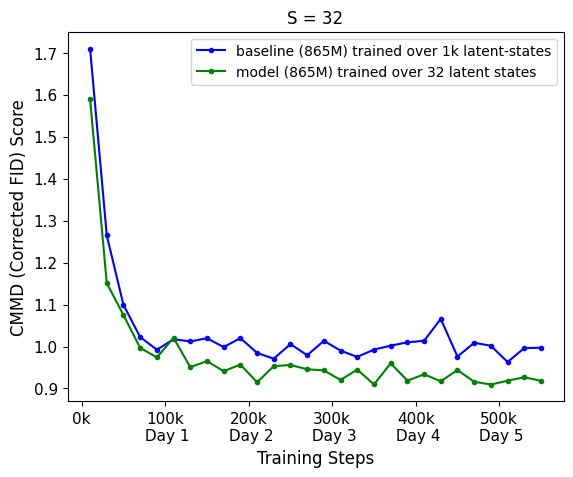}
        \end{minipage}
        \begin{minipage}[b]{0.24\textwidth}
            \includegraphics[width=\textwidth]{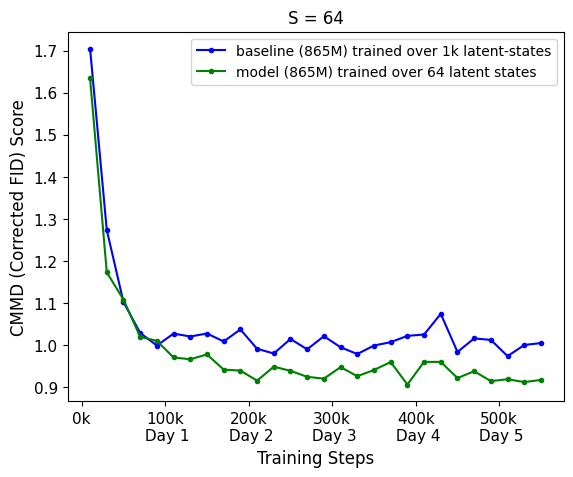}
        \end{minipage}
        \vspace{-0.4cm}
    \end{subfigure}    
    \begin{subfigure}{\textwidth}
        \centering
        \begin{minipage}[b]{0.24\textwidth}
            \includegraphics[width=\textwidth]{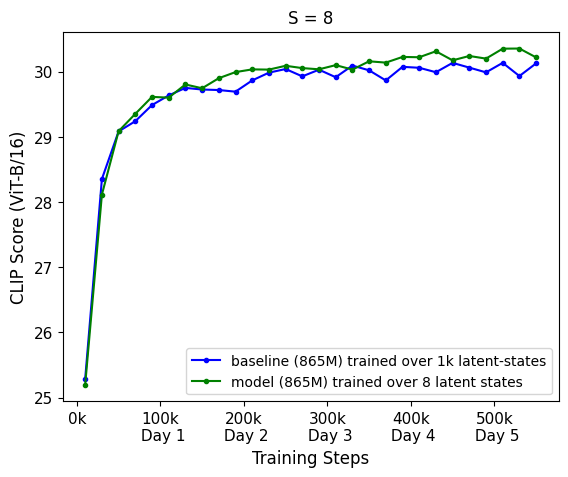}
        \end{minipage}
        \begin{minipage}[b]{0.24\textwidth}
            \includegraphics[width=\textwidth]{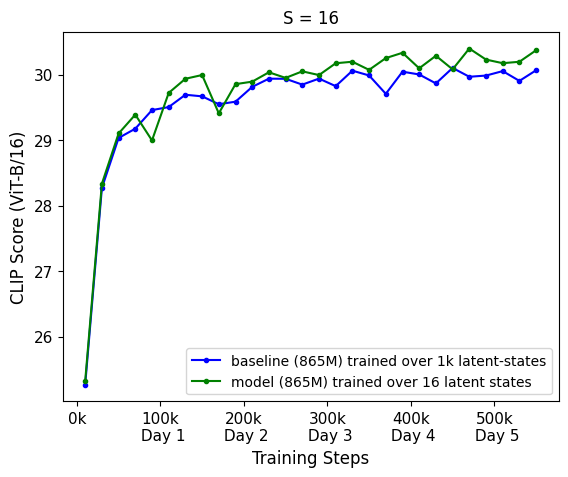}
        \end{minipage}
        \begin{minipage}[b]{0.24\textwidth}
            \includegraphics[width=\textwidth]{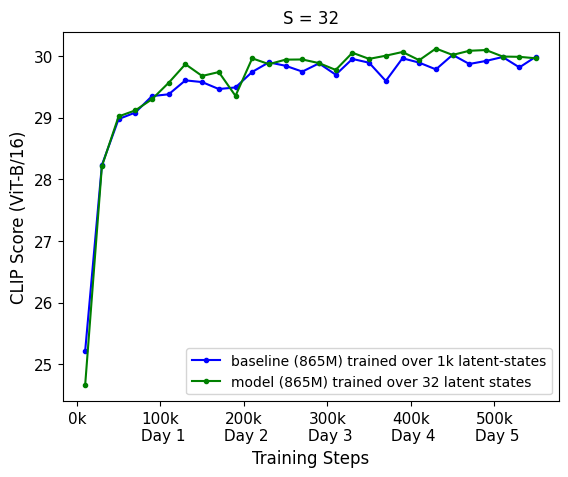}
        \end{minipage}
        \begin{minipage}[b]{0.24\textwidth}
            \includegraphics[width=\textwidth]{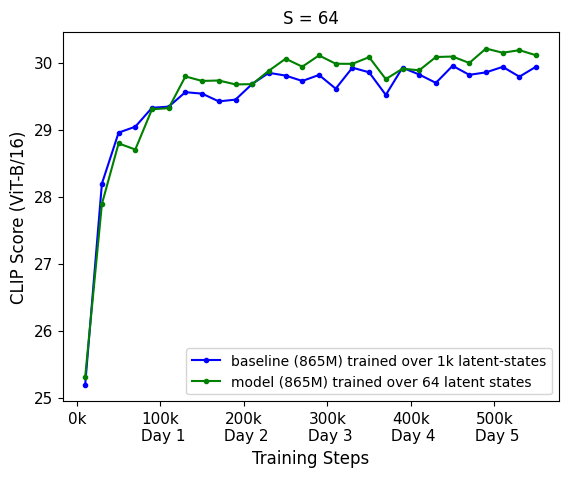}
        \end{minipage}
        \caption{Text-to-Image experiments with performance measured using TIFA, CMMD and CLIP scores. }\label{fig:diff_latent_t2i}
        \vspace{-0.2cm}
    \end{subfigure}
    
    \caption{Convergence of vanilla DDPM baseline model with 1000 latent-states (\ref{algo:ddpm_training}) vs the DDPM model trained over much fewer number of latent-states S with (\ref{algo:rev_training}). The baseline model is trained only once and with the same set of weights images are sampled for different number of inference steps. However for latter case, for different number of sampling steps S,  the models are trained from scratch only on the corresponding S latent-states. Surprisingly, it is feasible to effectively train diffusion models on much fewer latent-states!  
    } \label{fig:diff_latent}
    \vspace{-0.5cm}
\end{figure*}

%% file: sec/4_disentanglement.tex
\section{Complete Disentanglement in T-space} \label{sec:dis_t_space}
In the previous section we demonstrated that diffusion models can be effectively trained over a small number of latent states. Building on this, the goal of this section is to develop methods for faster training of diffusion models while also significantly improving their overall performance.



Contemporary diffusion models even with sufficient compute are slow to train using algorithm \ref{algo:ddpm_training}, due their low throughput (images processed per second). Naively increasing the batch size with more compute   does not necessarily lead to faster convergence as shown in \cite{scalability_agi}. 
Further to keep the run-time per iteration small with larger batch-sizes, compromises in terms of model size, token context length and image embeddings are required thus inevitably limiting model performance.

The goal of this paper is to build an approach which can overcome the limitation of low throughput. With this motivation, the key question  we aim to answer here is the following: Given \textbf{sufficient compute}, what is the minimum time in which a diffusion model can be trained?

To answer the above question, we build upon our findings in previous section of reducing the number of latent states required to train diffusion models. In particular, we push the limit of the number of latent-states to a \emph{minimum possible value of $1$}. \textbf{We define this as complete disentanglement in T-space.}  Here, a model (with parameters $\theta_{\tau_i}$) is trained to denoise only for a single SNR level corresponding to the latent-state $x_{\tau_i}$. Multiple such independently trained models corresponding to different SNR levels can then be combined to do inference using \eqref{eq:ddim_T_acc} to generate samples starting from pure Gaussian noise.  

In algorithm \ref{algo:limit_training}, we outline the training procedure for complete disentanglement in T-space. We train a separate model $\theta_t \; \forall \; t \;\in [\tau_1,\dots,\tau_\text{S}]$, where model $\theta_t$ denoises inputs only at a single SNR level corresponding to $t$. Importantly, note that here again corresponding $\alpha_t$'s are pre-computed using exactly the same schedule as in algorithm \ref{algo:ddpm_training} with $T=1,000$.

\begin{algorithm}
	\caption{Complete Disentangled Training (Ours)} \label{algo:limit_training}
    \begin{algorithmic}[1]
                \State \textbf{Input:} $[\tau_1,\dots,\tau_\text{S}], T$
                \For{ $t \leftarrow [ 1,\dots,T ]$}
                    \State Compute $\alpha_t$
                \EndFor
                \For{ $t \leftarrow [ \tau_1,\dots,\tau_\text{S} ]$}
                \Repeat
                \State ${x}_0 \sim q({x}_0),{\epsilon} \sim \mathcal{N}({0}, {I}  )$ 
                \State {\setlength{\abovedisplayskip}{-6pt}  \begin{alignat}{10}  \hspace{0.75cm} L_{\theta_t} = \Vert  {\epsilon} - {\epsilon}_{\theta_t}( \sqrt{{\alpha}_t}{x}_0 + \sqrt{ 1 - {\alpha}_t }{\epsilon}, \phi)   \Vert^{2}   \label{eq:distr_loss}
                \end{alignat}}
                \State $ \theta_t \leftarrow \theta_t - \eta \nabla_{\theta_t} L_{\theta_t}$
                \Until{converged}
                \EndFor
            \State \textbf{Output:} $[\theta_{\tau_1},\dots, \theta_{\tau_\text{S}}]$
	\end{algorithmic}     
\end{algorithm}

From a computational perspective, a \textbf{key aspect} of algorithm \ref{algo:limit_training} is that models corresponding to different SNR levels or latent-states can be trained independently and therefore in \textbf{parallel}. Depending on the number of latent states $\text{S}$, the maximum overall batch size can be $\text{S}$ times the batch size of a single latent-state model. This gives algorithm high flexibility to \emph{linearly} scale the overall throughput by training multiple single latent-state models concurrently.

\tikzstyle{mybox1} = [draw=cyan, fill=cyan!10, thick,
    rectangle, rounded corners, inner sep=8pt, inner ysep=8pt]

\begin{center}
\begin{tikzpicture}
\node [mybox1] (box){%
    \begin{minipage}{0.92\linewidth}
\textbf{
    \emph{Disentanglement in T-space using algorithm \ref{algo:limit_training} provides linear scaling in throughput (imgs/sec)!}
}
    \end{minipage}
};
\end{tikzpicture}%
\end{center}

In addition to the linear scaling of the throughput, algorithm \ref{algo:rev_training} is also \textbf{endowed with the following flexibilities}: 

\begin{enumerate}
    \item Compute nodes need not be located in the same data center or geographical region, i.e. different single latent-state models can be trained in completely different regions.
    \item Heterogeneity in type of compute nodes is now possible, thus
    allowing different single latent-state models to be trained on a mixture of available resources like H100/200s, A100s, V100s, etc. Also, different 
    models can be trained with different configurations including different batch-sizes, precision, gradient check-pointing etc. depending on the type of hardware used.
\end{enumerate}

These benefits are crucial in practice for increased cost-effectiveness and choice of computing resources.

At \textbf{inference}, we use these $\text{S}$ different single latent-state models to de-noise the input at their corresponding time-step using equation \eqref{eq:ddim_T_acc_distr}:
{\setlength{\abovedisplayskip}{-1pt}
\setlength{\belowdisplayskip}{0pt}
\begin{adjustwidth}{-0.25cm}{}
\begin{equation}
\footnotesize
{x}_{\tau_{i-1}} =  \bigg( \frac{{x}_{\tau_i} - \sqrt{1-\alpha_{\tau_i}} {\epsilon}_{\theta_{\tau_i}}({x}_{\tau_i},{\phi})}{\sqrt{\alpha_{\tau_i}/\alpha_{\tau_{i-1}}}   } \bigg) + \frac{ {\epsilon}_{\theta_{\tau_i}}({x}_{\tau_i},{\phi}) }{\sqrt{ (1-\alpha_{\tau_{i-1}})^{-1} }}
 \label{eq:ddim_T_acc_distr}
\end{equation}
\end{adjustwidth}
}
where $\epsilon_\theta(x_t,t)$ in \eqref{eq:ddim_T_acc} is replaced with $\epsilon_{\theta_{\tau_i}}(x_t,\phi)$ to derive \eqref{eq:ddim_T_acc_distr}.  
Note that since the total number of sampling steps required remains the same  as a typical diffusion model i.e.  $\text{S}$, therefore the inference time (or the overall throughput) also remains the same. For the case when all the single latent-state models do not fit in a single GPU, different models can be split across multiple GPUs. The additional inter-GPU communication overhead is marginal ($\le 2\%$, see table \ref{tab:throughput_S_32}) because of
  very high communication bandwidths of the order of  $10^2-10^3$ GB/sec possible due to NVLink \cite{nvlink_technology}. 



\subsection{Experiments-II (Complete Disentanglement)}
In this section we benchmark completely disentangled model (\ref{algo:limit_training})  against the  baseline model (\ref{algo:ddpm_training}) i.e. vanilla DDPM. 
Similar to section \ref{sec:exp1}, we here again evaluate on two different setups i.e. ImageNet and Text-to-image. For both data-sets, the baseline models and the evaluation metrics remains the same as the ones used in section \ref{sec:exp1}.

For the disentangled model, each single latent-state model is trained independently, and follows the same setup as the baseline in terms of the U-Net network architecture, dataset pre-processing, batch-size  and learning rate etc.
For instance, when $\text{S} = 32$, the disentangled model comprises of $32$ single-latent state models   corresponding to $t \in \{ 31,62, 93\dots, 936,967,999 \}$. Importantly, here each model only learns to denoise input images at a single SNR level.

 After training each of the individual  single latent-state models, we combine these $32$ models to sample images using equation (\ref{eq:ddim_T_acc_distr}).
 The convergence on ImageNet dataset measured using FID and IS is show in fig. \ref{fig:base_vs_limit_cin}. Similarly, fig. \ref{fig:base_vs_limit_t2i} shows the convergence on text-to-image task using CLIP, TIFA and CMMD scores at $256$ resolution. Importantly, the disentangled model for $\text{S}= 16,32,64$ provides much faster convergence on both the datasets. In particular, for text-to-image task, disentangled model after only 1 day of training matches the performance of the baseline model trained over five days. 

  
 Next using the disentangled model we want to generate high quality samples with only one day of total training time. To this end, we train all the 32 single latent state models for 85k iterations at 256 resolution and then train for another 15k iterations at 512 resolution, i.e. maintaining a strict total budget of 1 day of training time. We compare against the baseline model which is trained for 550k iterations at 256 resolution and then for another 50k iterations at 512 resolution, i.e a total training time of 5 days.
 A comparison of generated samples is shown in figure \ref{fig:teaser}.  From figure \ref{fig:teaser}, we note that the samples generated from the disentangled model after 1 day of training match the visual quality of the samples generated  after 5 days of training. More samples are provided in Appendix.

Finally, table \ref{tab:throughput_S_32} compares the sampling throughput for the baseline with $32$ sampling steps and the disentangled model composed of $32$ single latent-state models. We note that the throughput for both the models stabilizes for batch-sizes $\ge 20$, and the reduction (\% diff.) is marginal and well below 2\%.   
\begin{table}[]
\centering
\resizebox{\columnwidth}{!}{%
\begin{tabular}{@{}lcccccccc@{}}
\toprule
Batch-size    & 8    & 12   & 16   & 20   & 24   & 28   & 32   & 36   \\ \midrule
Baseline      & 1.93 & 1.99 & 2.04 & 2.07 & 2.08 & 2.09 & 2.11 & 2.11 \\
Disentangled & 1.85 & 1.94 & 1.99 & 2.04 & 2.05 & 2.06 & 2.08 & 2.08 \\ \midrule
\% Diff.       & 4.15 & 2.51 & 2.45 & \bf{1.45} & \bf{1.44} & \bf{1.44} & \bf{1.42} & \bf{1.42} \\ \bottomrule
\end{tabular}%
}
\caption{Inference throughput (images/sec) comparison between baseline (\ref{eq:ddim_T_acc}) and disentangled model (\ref{eq:ddim_T_acc_distr}) for S =32 sampling steps for T2I task.}
\label{tab:throughput_S_32}
\vspace{-0.5cm}
\end{table}

\subsubsection*{Why disentanglement provides higher convergence?} Obviously, large overall model size (x S) is one of the  reasons. However, note that training a vanilla model of comparative size with same compute may not be even feasible to begin with or will be plagued with extremely low throughput (images/day). 
On the contrary, a major benefit of disentanglement is the feasibility of very high through-puts while maintaining large overall model sizes or increased network capacity. Since S different models (of varied sizes) inherently can be trained in a distributed manner (across different geographical regions) in parallel, therefore the throughput is also scaled by a factor of S. For context, the vanilla baseline model consumes a meager $100$M images per day, while the disentangled model for $\text{S}=32$ has a throughput of $3.2$ billion images/day!

\input{sec/fig_disent}

\subsection{How to reduce compute requirements?}
Training S different single latent-state models can be challenging with limited compute. This can be address to a certain extent by exploiting the fact that all single latent-state models do not require equal amount of training. The hardness of the underlying optimization problem, i.e. to learn the parameters $\theta_t$ in (\ref{algo:limit_training}) varies significantly with the value of $\alpha_t$ or the SNR level \cite{trunc_ddp}. Using this aspect, the compute required can reduced by as much as 50\% as discussed in detail in Appendix. 
A complementary approach could involve employing varied model sizes for different latent states. In particular, using smaller models for
latent-states corresponding low SNR.
\section{Related Work} 
 Prior works like \cite{trunc_ddp, ddgan}  couple 
 diffusion models with other generative models like GANs resulting reduced the number of latent-states. However, 
 due to this coupling the effect of altering the number of latent-states is not clear and properly understood.

There also have been numerous efforts to improve the convergence of diffusion models. These include using a different weighting strategy in the training loss \cite{a_closer_look, min_snr, perception_prioritized} or a non-uniform distribution for time-steps \cite{eluc_design_space,  clts,  a_closer_look, non_unf_t_samp_faster_diffusion, beta_tuned_t_diffusion}. Efforts in the direction of network architecture and masking include \cite{masket_dit_strong_synthesizer, pyramidal_ddpm, patch_diffusion, mask_dit, switch_Dit}. Other related works would include \cite{ediff-i,ernie} where the goal is to improve overall performance or sample quality by fine-tuning an already trained base model into multiple experts.
\section{Conclusion}
In this paper, we challenge the fundamental assumption that large number of latent-states is required to train diffusion models. We show that by careful selection of the noise schedule parameters,
models with much fewer number of latent states can be effectively trained  without any loss in performance.
Interestingly, it is feasible to reduce the number of latent-states to one resulting in single latent-state models or complete disentanglement is T-space. Our experiments on two different datasets show that by combining multiple single latent-state models, speed-ups of upto $6\times$ can be achieved 
along with much superior performance. Further, there are important additional benefits such as the flexibility to train across different geographical regions and hardware.

%% file: sec/fig_disent.tex
\begin{figure*}[htbp]
    \centering
    \begin{subfigure}{\textwidth}
        \centering
        \begin{minipage}[b]{0.24\textwidth}
            \includegraphics[width=\textwidth]{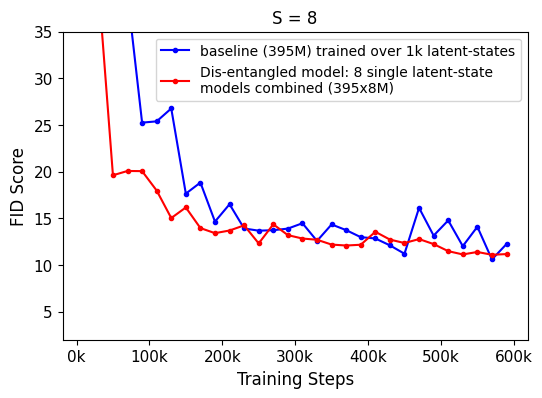}
        \end{minipage}
        \begin{minipage}[b]{0.24\textwidth}
            \includegraphics[width=\textwidth]{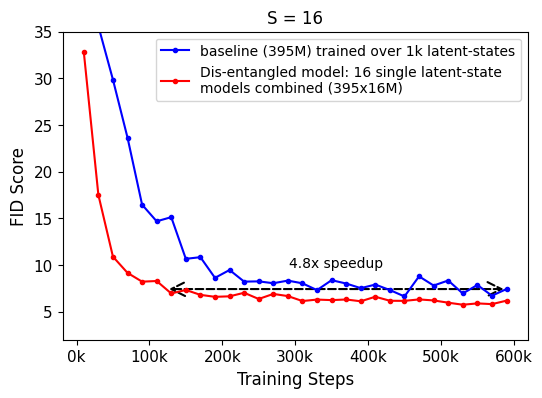}
        \end{minipage}
        \begin{minipage}[b]{0.24\textwidth}
            \includegraphics[width=\textwidth]{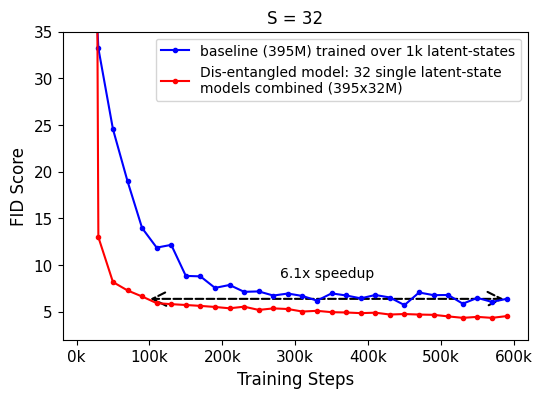}
        \end{minipage}
        \begin{minipage}[b]{0.24\textwidth}
            \includegraphics[width=\textwidth]{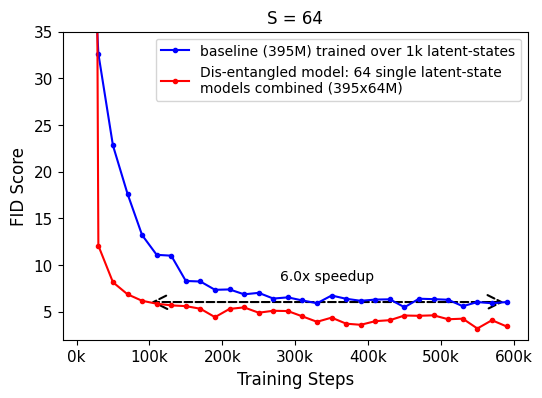}
        \end{minipage}
    \end{subfigure}    
    \vspace{4mm}
    \begin{subfigure}{\textwidth}
        \centering
        \begin{minipage}[b]{0.24\textwidth}
            \includegraphics[width=\textwidth]{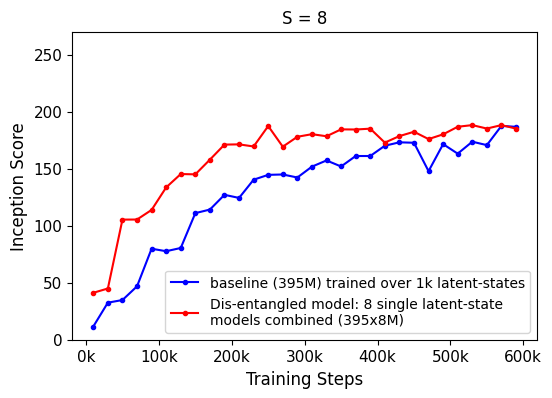}
        \end{minipage}
        \begin{minipage}[b]{0.24\textwidth}
            \includegraphics[width=\textwidth]{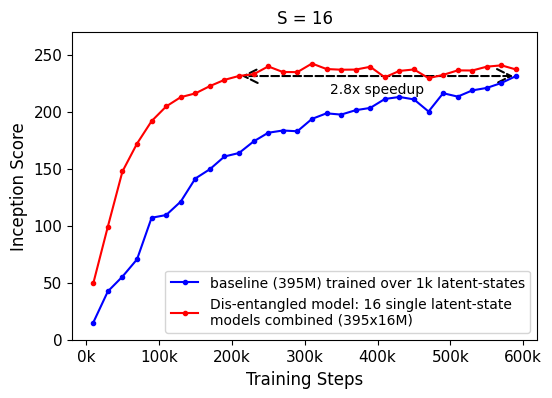}
        \end{minipage}
        \begin{minipage}[b]{0.24\textwidth}
            \includegraphics[width=\textwidth]{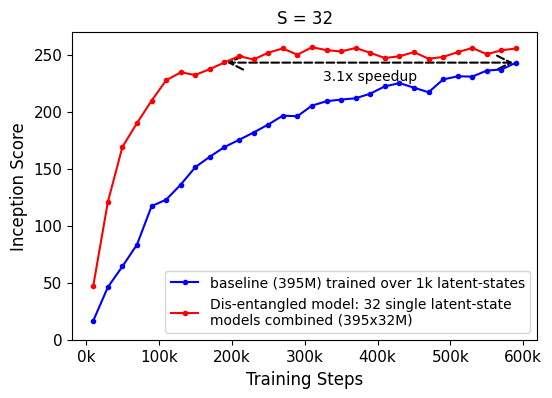}
        \end{minipage}
        \begin{minipage}[b]{0.24\textwidth}
            \includegraphics[width=\textwidth]{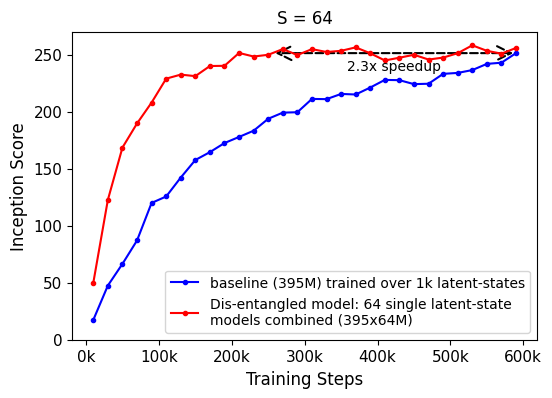}
        \end{minipage}
        \caption{ImageNet: Disentangled model provides 4-6$\times$ faster convergence measured using FID and Inception scores.}\label{fig:base_vs_limit_cin}
        \vspace{-0.18cm}
    \end{subfigure}
    \begin{subfigure}{\textwidth}
        \centering
        \begin{minipage}[b]{0.24\textwidth}
            \includegraphics[width=\textwidth]{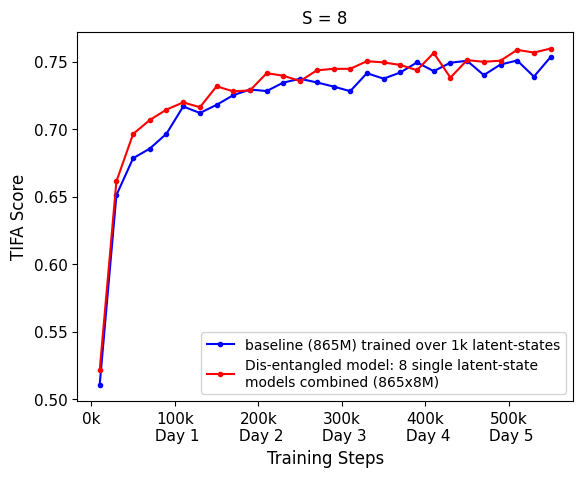}
        \end{minipage}
        \begin{minipage}[b]{0.24\textwidth}
            \includegraphics[width=\textwidth]{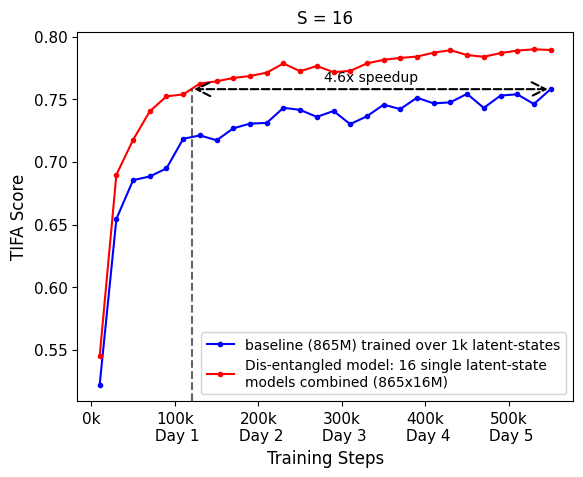}
        \end{minipage}
        \begin{minipage}[b]{0.24\textwidth}
            \includegraphics[width=\textwidth]{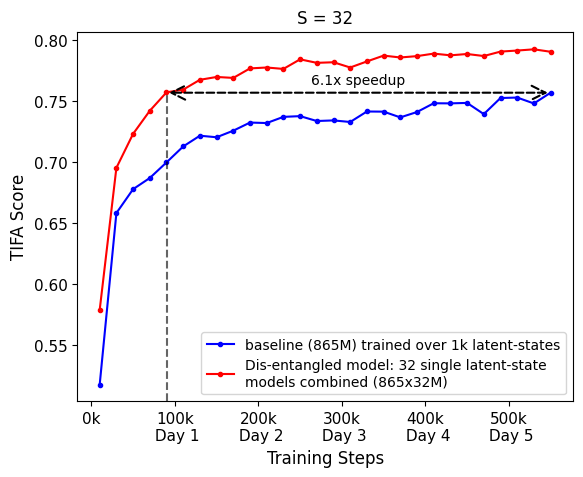}
        \end{minipage}
        \begin{minipage}[b]{0.24\textwidth}
            \includegraphics[width=\textwidth]{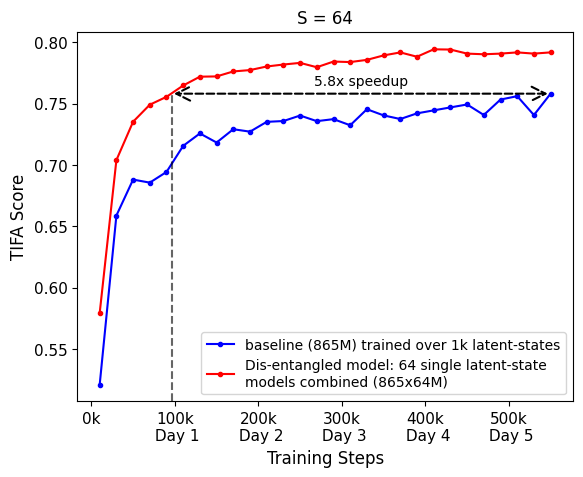}
        \end{minipage}
    \end{subfigure}
    \begin{subfigure}{\textwidth}
        \centering
        \begin{minipage}[b]{0.24\textwidth}
            \includegraphics[width=\textwidth]{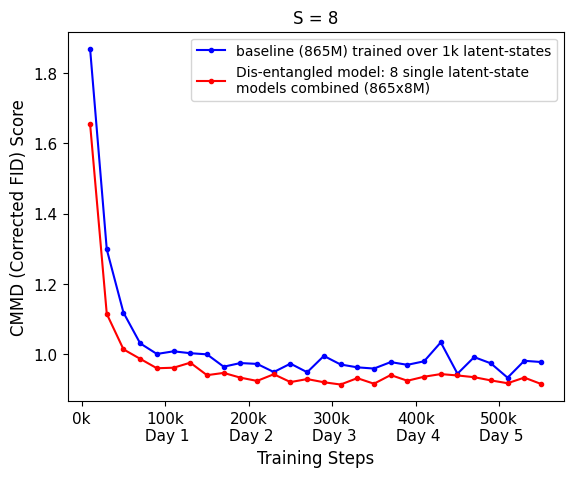}
        \end{minipage}
        \begin{minipage}[b]{0.24\textwidth}
            \includegraphics[width=\textwidth]{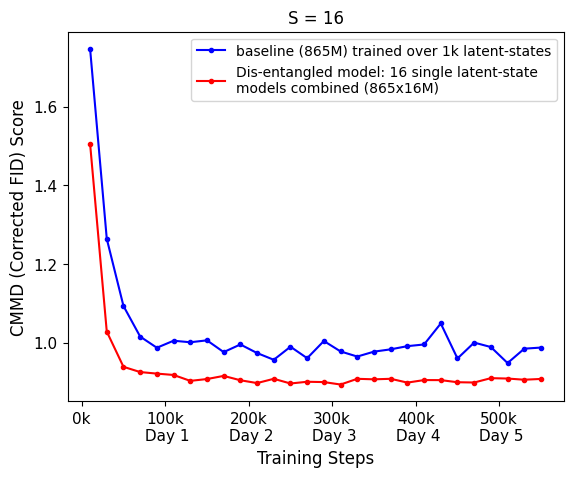}
        \end{minipage}
        \begin{minipage}[b]{0.24\textwidth}
            \includegraphics[width=\textwidth]{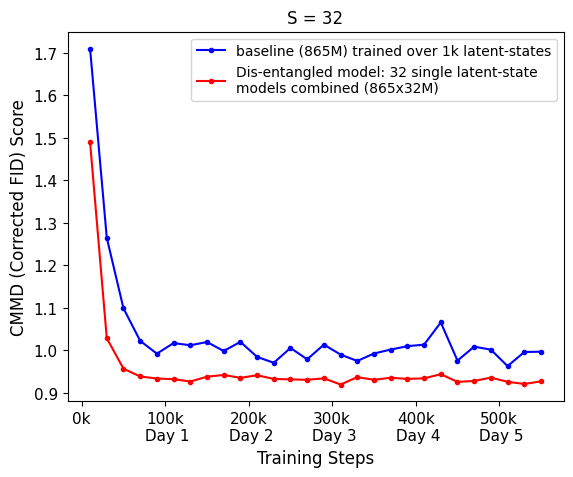}
        \end{minipage}
        \begin{minipage}[b]{0.24\textwidth}
            \includegraphics[width=\textwidth]{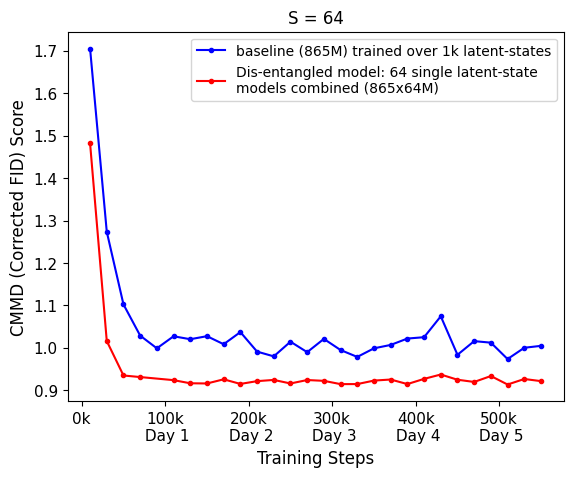}
        \end{minipage}
    \end{subfigure}    
    \begin{subfigure}{\textwidth}
        \centering
        \begin{minipage}[b]{0.24\textwidth}
            \includegraphics[width=\textwidth]{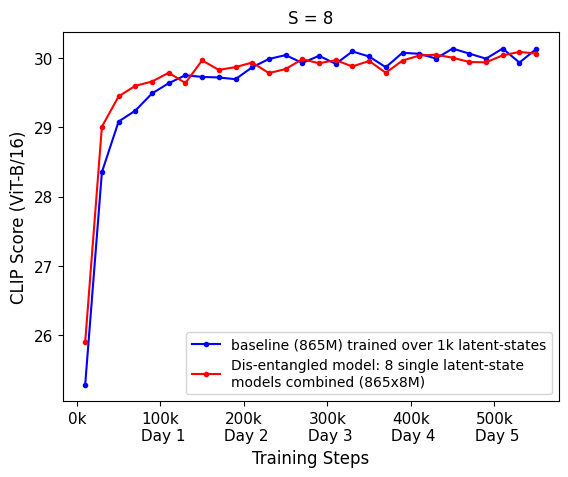}
        \end{minipage}
        \begin{minipage}[b]{0.24\textwidth}
            \includegraphics[width=\textwidth]{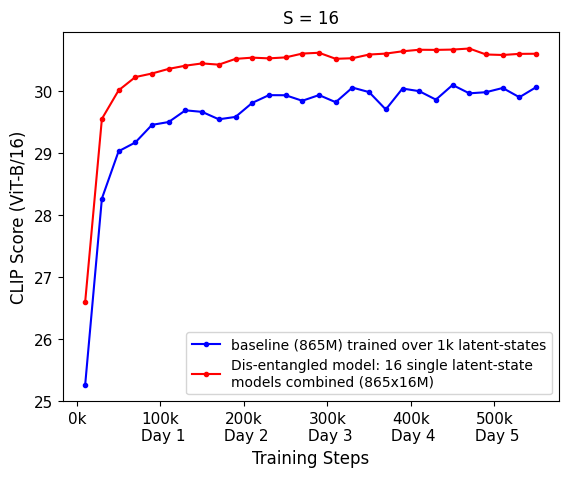}
        \end{minipage}
        \begin{minipage}[b]{0.24\textwidth}
            \includegraphics[width=\textwidth]{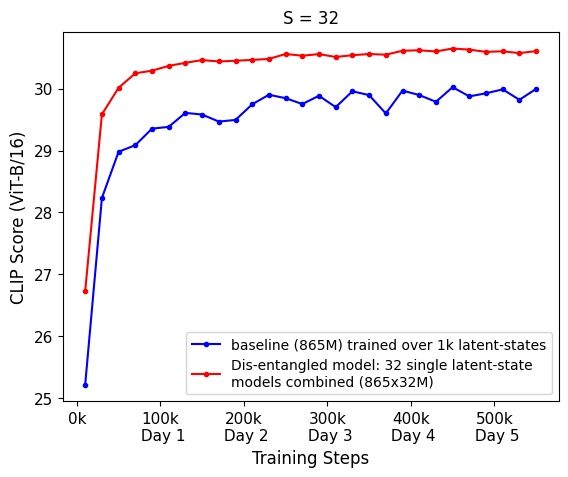}
        \end{minipage}
        \begin{minipage}[b]{0.24\textwidth}
            \includegraphics[width=\textwidth]{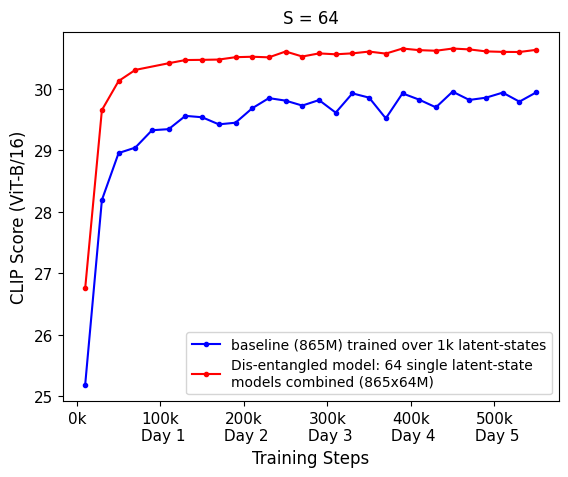}
        \end{minipage}
        \caption{Text-to-image: Disentangled model trained only for one day can match the performance of the baseline model trained over several days.}\label{fig:base_vs_limit_t2i}
    \end{subfigure}
   \vspace{-15pt}
    \caption{ Convergence of vanilla DDPM baseline model with 1000 latent-states  (\ref{algo:ddpm_training}) vs disentangled model (\ref{algo:limit_training}). The baseline model is trained only once and with the same set of weights images are sampled for different number of inference steps. For the disentangled model, during inference for each latent-state a separate single-latent state model trained  only on that corresponding latent-state is used. 
    }
    \label{fig:base_vs_limit}
\end{figure*}

%% file: sec/X_suppl.tex
\onecolumn
\clearpage
\section{\phantom{mmmmmmmmmmmmmmmmmmmmm} Appendix\\
}

\subsection{Sampling Algorithms:}

\begin{algorithm}
	\caption{DDPM (Markovian) Sampling  \cite{ho_ddpm} } \label{algo:ddpm_inference}
	\begin{algorithmic}[1]
                \For{$t= T,\dots,1$}
                \State ${x}_{T} \sim \mathcal{N}({0},{I})$
                \State ${\epsilon} \sim \mathcal{N}({0},{I}) \text{ if } t>1, \text{ else } {\epsilon} = {0}$
                \State   {\setlength{\abovedisplayskip}{-5pt}
 \begin{alignat}{10}
               \hspace{0.5cm} {x}_{t-1} =& \sqrt{\frac{\alpha_{t-1}}{\alpha_t}} \bigg({x}_t - \frac{\alpha_{t-1} -\alpha_{t} }{\alpha_{t-1} \sqrt{1-\alpha_t}} {\epsilon}_{\theta}({x}_t,t) \bigg) + \notag \\
                &\sqrt{\frac{1-\alpha_{t-1}}{1-\alpha_t}\bigg( 1- \frac{\alpha_t}{\alpha_{t-1}} \bigg)      } \; \cdot\,{\epsilon}  \hspace{0.3cm}     \label{eq:supp_ddpm_infer}             
                \end{alignat}}
                
                \EndFor
                \State \Return ${x}_0$
	\end{algorithmic} 
\end{algorithm}

\begin{algorithm}
	\caption{Generalised (Non-Markovian) Sampling \cite{ddim} } \label{algo:ddim_inference}
	\begin{algorithmic}[1]
                \For{$t= T,\dots,1$}
                \State ${x}_{T} \sim \mathcal{N}({0},{I})$
                \State ${\epsilon} \sim \mathcal{N}({0},{I}) \text{ if } t>1, \text{ else } {\epsilon} = {0}$
                \State {\setlength{\abovedisplayskip}{-5pt}  \begin{alignat}{10}
                \hspace{0.225cm} {x}_{t-1} =\;  & \sqrt{\alpha_{t-1}}\bigg( \frac{{x}_t - \sqrt{1-\alpha_t} {\epsilon}_{\theta}({x}_t,t)}{\sqrt{\alpha_t}} \bigg) +  \notag  \\
                & \sqrt{ 1-\alpha_{t-1} -\sigma_t^2}\cdot {\epsilon}_{\theta}({x}_t,t) + \sigma_t {\epsilon} \label{eq:supp_ddim_T}
                \end{alignat} } 
                \EndFor
                \State \Return ${x}_0$
	\end{algorithmic} 
\end{algorithm}

Note that when
\begin{equation}
\sigma_t = \sqrt{(1-\alpha_{t-1})/ (1-\alpha_t)}\sqrt{1-\alpha_t/\alpha_{t-1}} \notag
\end{equation}
\eqref{eq:supp_ddim_T} becomes identical to \eqref{eq:supp_ddpm_infer}, \ie, we recover the Markovian DDPM sampling \eqref{eq:supp_ddpm_infer}.

\clearpage
\subsection{How to reduce compute requirements?}

An important aspect of training single latent-state models as a part of disentangled training is that each individual model is trained to de-noise the input at a particular SNR level through the loss defined in equation \eqref{eq:distr_loss} in algorithm \ref{algo:limit_training}.  Importantly, 
the hardness of the underlying optimization problem, i.e. to learn the parameters $\theta_t$ varies significantly with the value of $\alpha_t$ or the SNR level \cite{trunc_ddp}.
We can exploit this characteristic to reduce the computational requirements of algorithm \ref{algo:limit_training}.
In practice, this translates to training different single-latent state models
for different number of iterations depending upon the SNR level of the latent state.

\begin{figure}[h]
    \centering
    \includegraphics[width=0.8\linewidth]{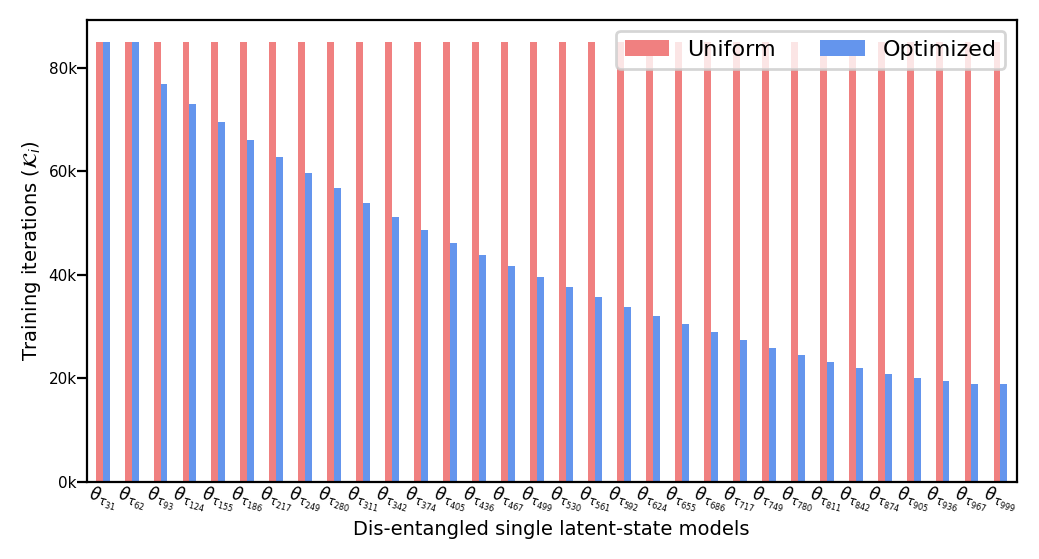}
    \vspace{-0.2cm}
    \caption{Two different disentangled models for $\text{S}=32$ with similar performance in terms of TIFA,CLIP and CMMD scores. In the uniform case (in red) each single-latent state model is trained for 85k iterations while in the optimized case (in blue) the number of training iterations decrease in proportion to the SNR levels of the individual latent-states. The optimized disentangled model (blue) requires almost $50\%$ less compute than uniform (red)!}
       
    \label{fig:optimal_distr}
\end{figure}

Figure \ref{fig:optimal_distr} shows two different distributions of training iterations across various $32$ latent-states. The resulting two different disentangled models achieve very similar TIFA, CLIP and CMMD scores. Importantly, the optimized distribution requires $50$\% less compute
as the individual models can be trained in batches on half the number GPUs. A complementary approach could involve employing varied model sizes for different latent states. In particular, using smaller models for
latent-states corresponding to large t or low SNR.

\subsection{A note on contemporary Diffusion models}
The following facts prevent faster training and restrict diffusion models from realising their full potential.

\begin{enumerate}    
\item \textbf{Large-Scale Datasets:} The datasets used for training foundation models typically contain 100 million to more than 5 billion image-text pairs to provide broad coverage of real-world concepts \cite{laion_5b}.

\item \textbf{Large Model Sizes:} Large models are required to faithfully learn representations for real-world concepts, ranging from 865 million parameters in SD-2.1 to 2.6 billion in SD-XL to 12 billion parameters in more recent models like FLUX.

\item \textbf{Limited VRAM:} Due to large model sizes, distributed training techniques like Distributed Data Parallel (DDP) or Fully Sharded Data Parallel (FSDP) are required, which necessitates inter-node communication. This means all nodes need to be in the same data center, preferably in the same network spine to minimize latency. 

\item \textbf{Imperfect Image Embeddings:} For computational efficiency training is done in the latent space using VAEs \cite{stable_diff} instead of pixel space. Since VAEs provide lossy compression, therefore the final image quality depends on well the VAE itself has beeen trained. Using larger VAEs with more channels improve quality \cite{sdxl,scalability_agi} at the cost of increasing training time.

\item \textbf{Limited Token Length of Text Embeddings:} The cross-attention layer for conditioning on text embedding is computationally expensive. Consequently, text embeddings with limited maximum token length are used. This constraint limits the model's ability to effectively process longer prompts or learn from image-text pairs with high compositional complexity.

\item \textbf{Limitations on larger batch sizes:} Typical training configurations employ batch sizes ranging from 512 to 2,048 with a learning rate of around $10^{-4}$ per iteration. Given the massive size of the datasets as discussed previously, the number of training iterations often run into millions for pre-training to effectively cover all the modes in the underlying data-distribution. To reduce the total number of training iterations, a naively obvious workaround would be to train with larger batch sizes. However, this approach faces some key challenges:
\begin{enumerate}
    \item With larger batch sizes training dynamics need to be stabilized by correctly scaling the learning rate and numerical issues may arise since training is typically done with half/mixed precision (fp16 or bf16) due to limited VRAM. 
    \item More importantly, \emph{increasing the batch size does not necessarily lead to faster convergence} as shown in \cite{scalability_agi}. 
\end{enumerate}
Due to diminishing returns of larger batch sizes in terms of convergence, simply increasing the batch size with more compute is not a straightforward solution to reducing the total number of required training iterations.

\end{enumerate}

\clearpage
\subsection{Sampling with fewer latent state models}
\begin{figure}[h]
    \centering
    \begin{subfigure}{0.4\textwidth}
        \includegraphics[width=\textwidth]{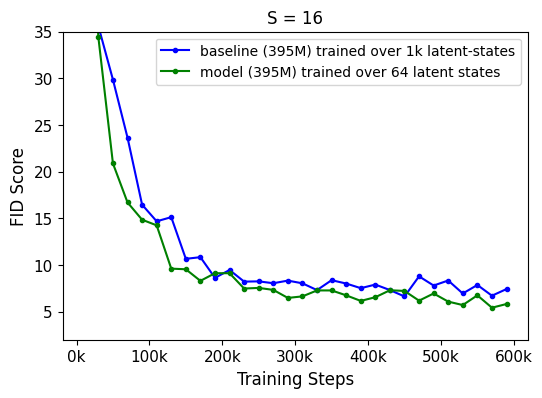}
        \label{fig:figure1}
    \end{subfigure}
    \hspace{1cm}
    \begin{subfigure}{0.4\textwidth}
        \includegraphics[width=\textwidth]{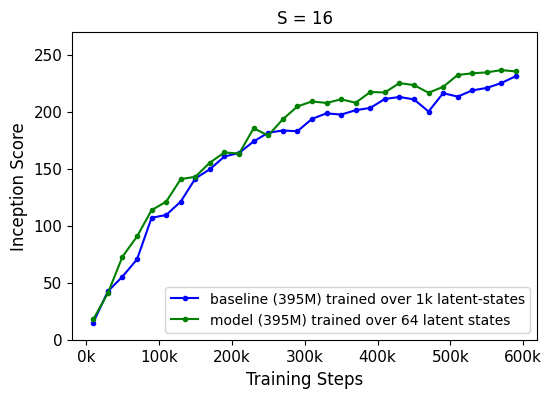}
        \label{fig:figure2}
    \end{subfigure}
    \vspace{-0.5cm}
    \caption{FID and IS scores for baseline model trained over 1k latent states and fewer-latent states model trained over 64 latent-states. But sampling is done for 16 steps for both models. }
    \label{fig:single64_S_16}
\end{figure}

\begin{figure}[h]
    \centering
    \begin{subfigure}{0.4\textwidth}
        \includegraphics[width=\textwidth]{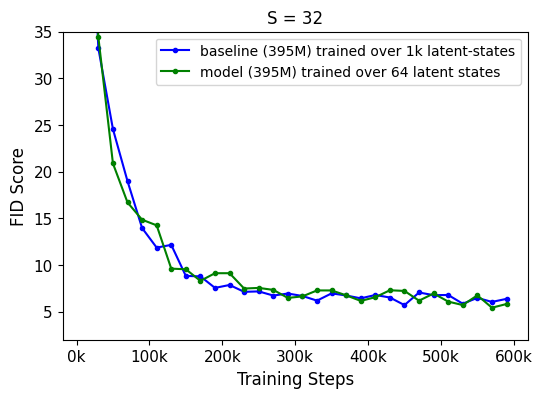}
        \label{fig:figure1}
    \end{subfigure}
    \hspace{1cm}
    \begin{subfigure}{0.4\textwidth}
        \includegraphics[width=\textwidth]{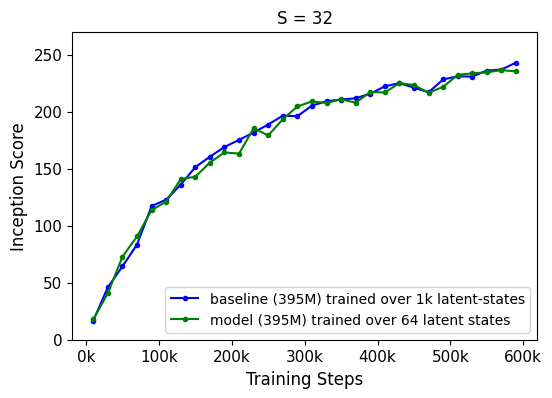}
        \label{fig:figure2}
    \end{subfigure}
    \vspace{-0.5cm}
    \caption{FID and IS scores for baseline model trained over 1k latent states and fewer-latent states model trained over 64 latent-states. But sampling is done for 32 steps for both models. }
    \label{fig:single64_S_32}
\end{figure}

\begin{figure}[h]
    \centering
    \begin{subfigure}{0.4\textwidth}
        \includegraphics[width=\textwidth]{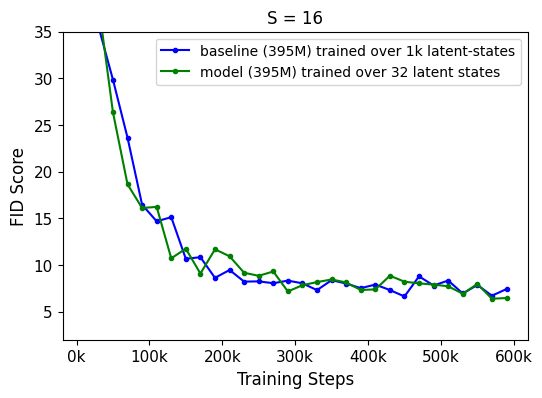}
        \label{fig:figure1}
    \end{subfigure}
    \hspace{1cm}
    \begin{subfigure}{0.4\textwidth}
        \includegraphics[width=\textwidth]{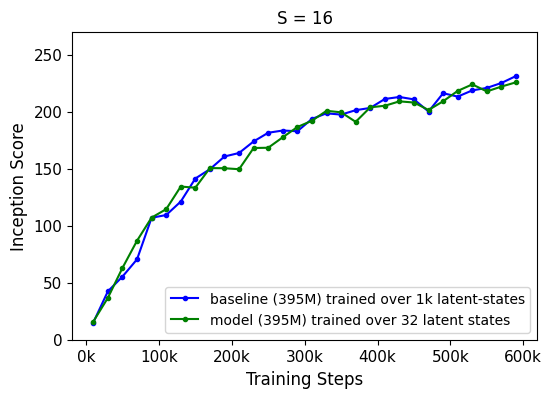}
        \label{fig:figure2}
    \end{subfigure}
    \vspace{-0.5cm}
    \caption{FID and IS scores for baseline model trained over 1k latent states and fewer-latent states model trained over 32 latent-states. But sampling is done for 16 steps for both models. }
    \label{fig:single32_S_16}
\end{figure}

A fewer latent-state model 
can be used to generate samples by sampling for an even lower number of sampling steps. For a $64$ latent-state model, we can generate samples for  $16$ or $32$ as show by the convergence plots in figure \ref{fig:single64_S_16} and \ref{fig:single64_S_32} respectively.
Similarly a fewer latent-states model trained over $32$ states can be used for sampling at $16$ steps as shown in figure \ref{fig:single32_S_16}.

\newpage
\def\imagelistbaseline{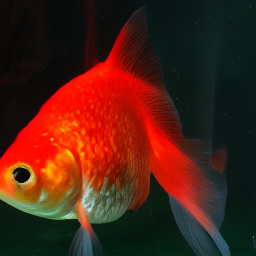, 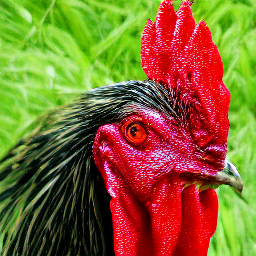,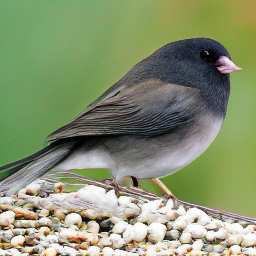,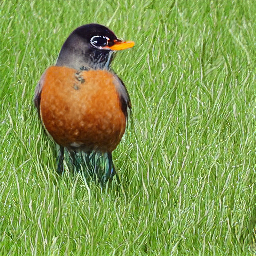,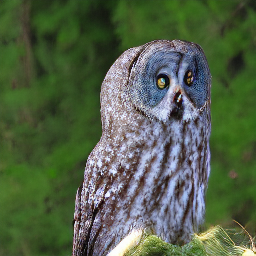,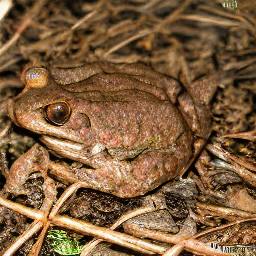,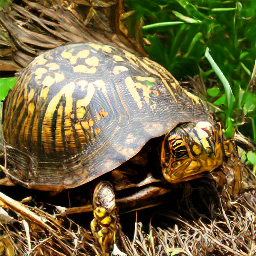, 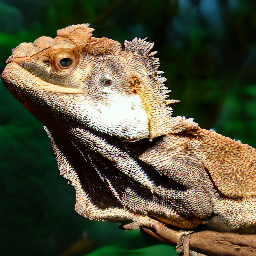, 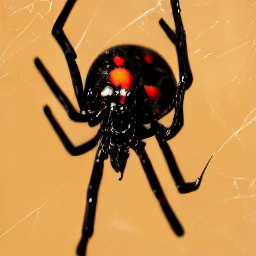,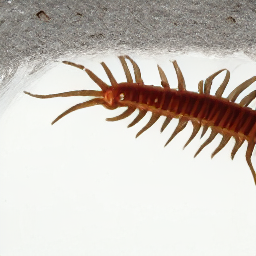, 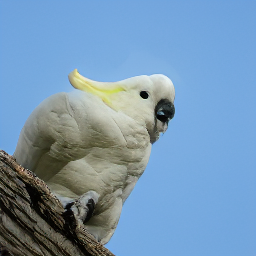, 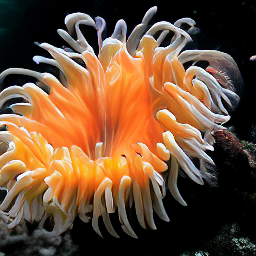,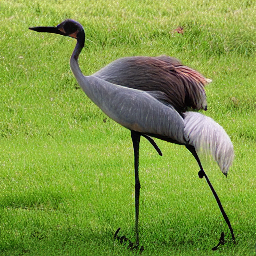, 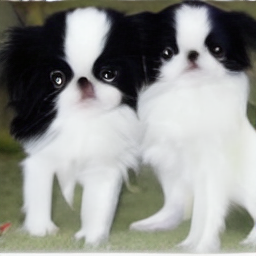, 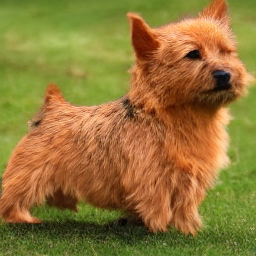,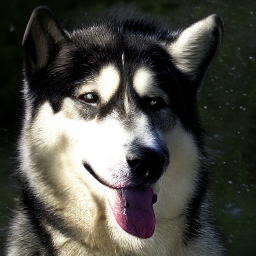, 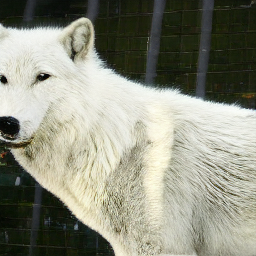, 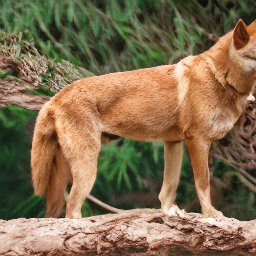,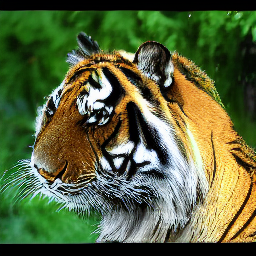, 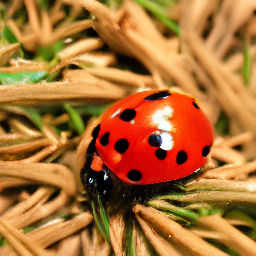, 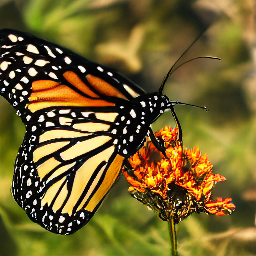,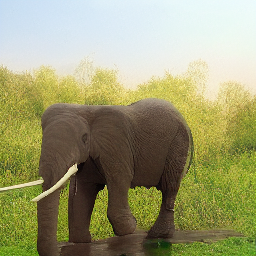, 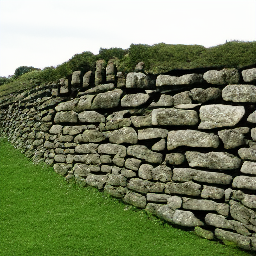, 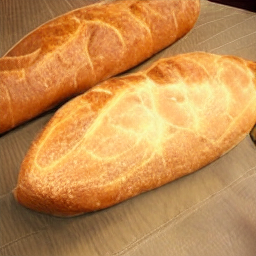,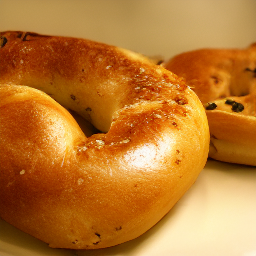, 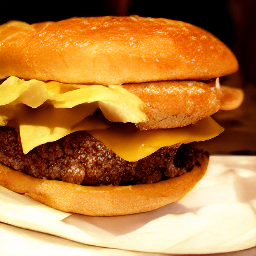, 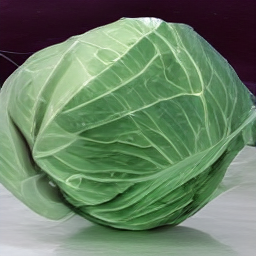,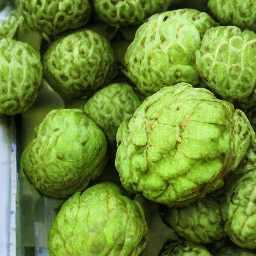, 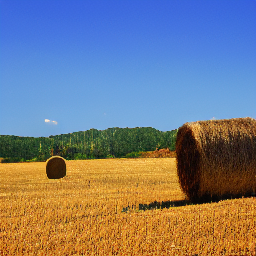, 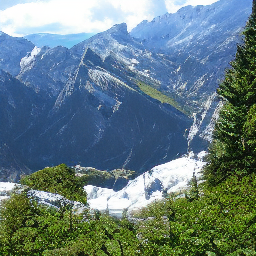,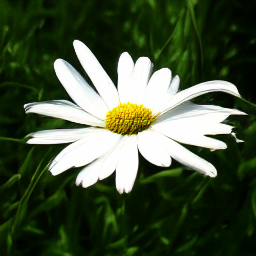,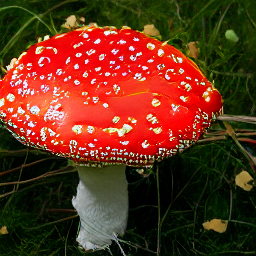}

\def\imagelistdisen{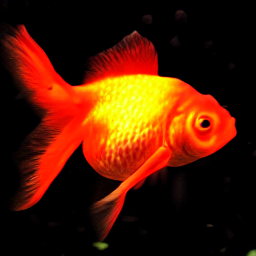, 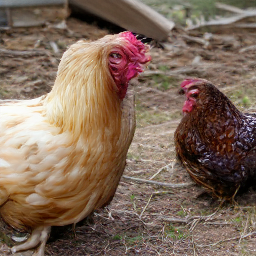, 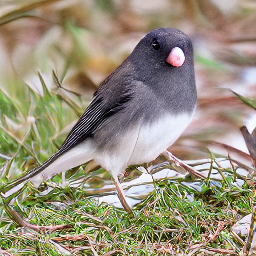, 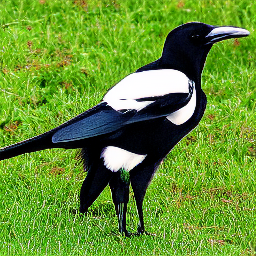, 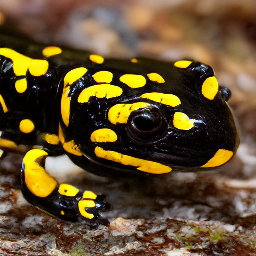, 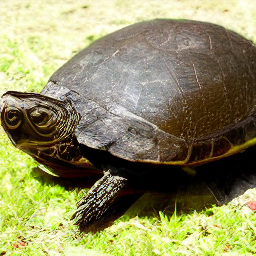,  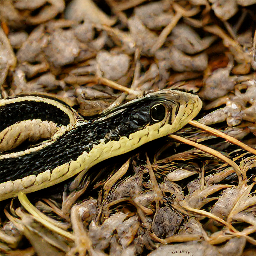, 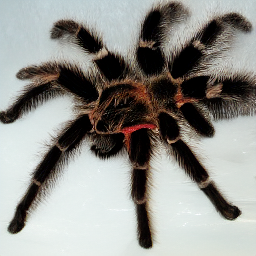, 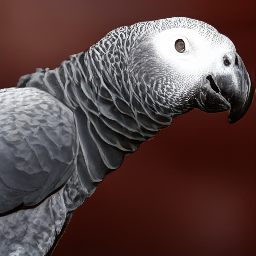, 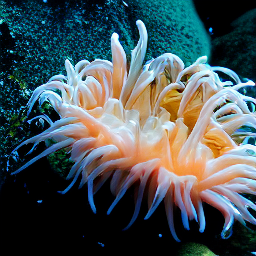, 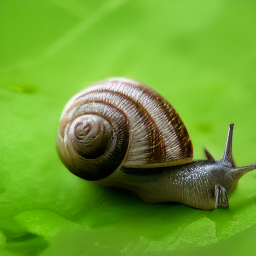, 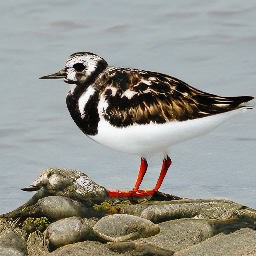, 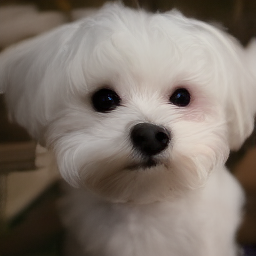, 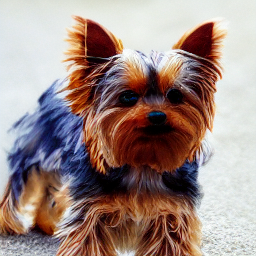, 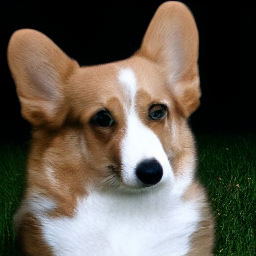, 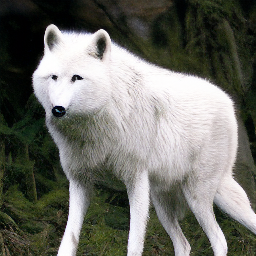, 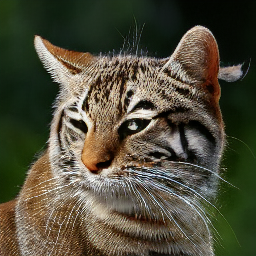, 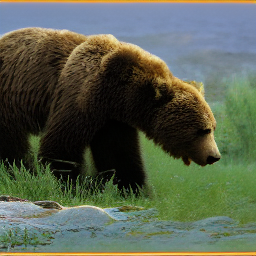, 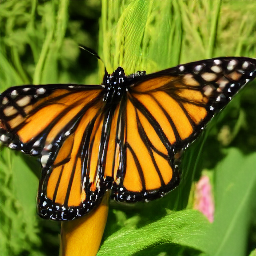, 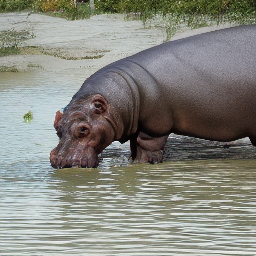, 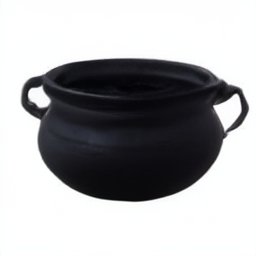, 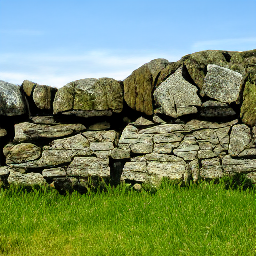, 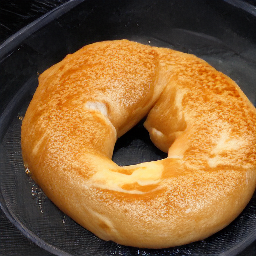, 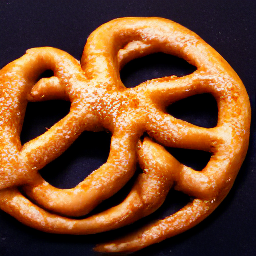, 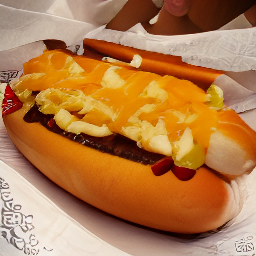, 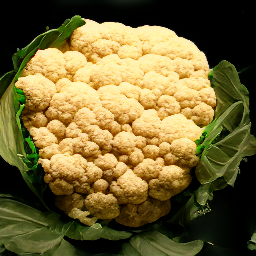, 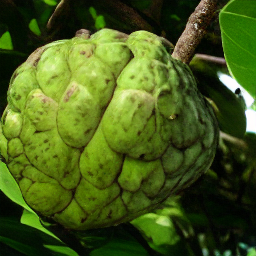, 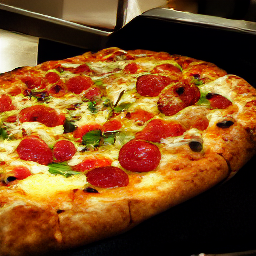, 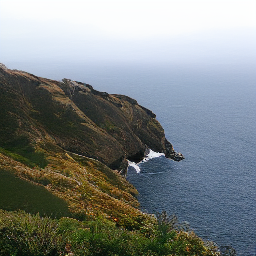, 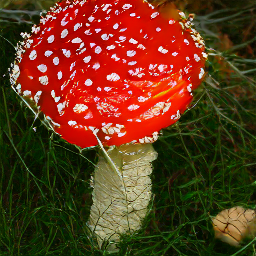, 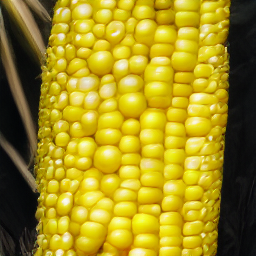,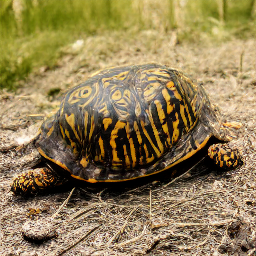,}

\def\imagelistsingle{
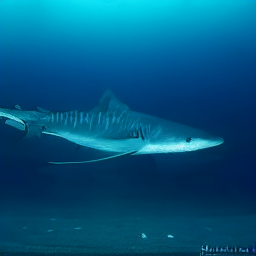, 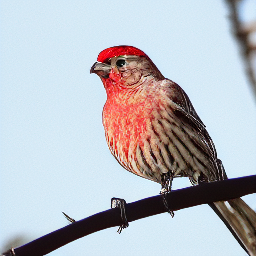, 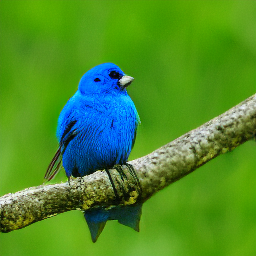, 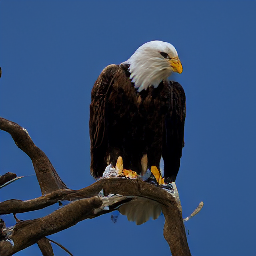, 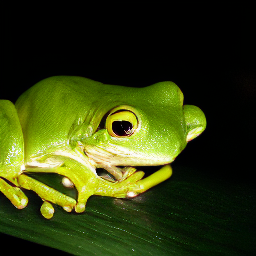, 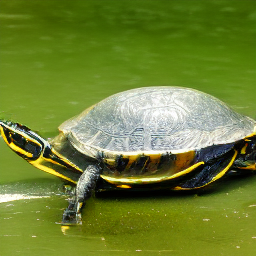, 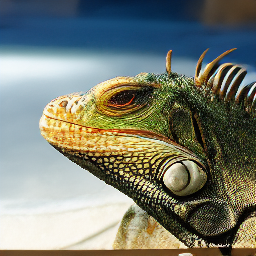, 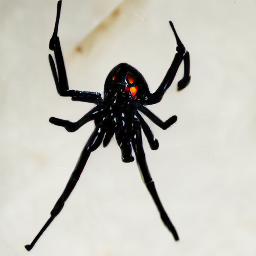, 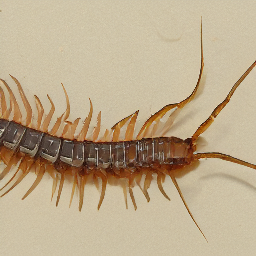, 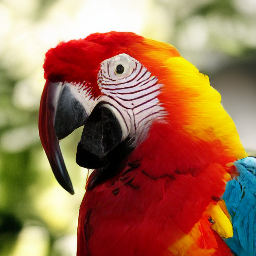, 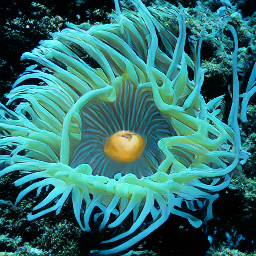, 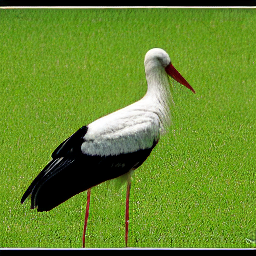, 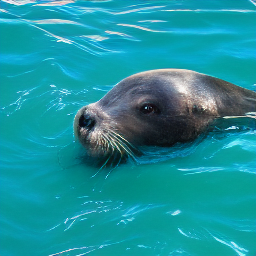, 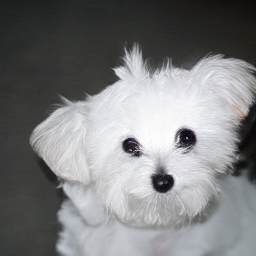, 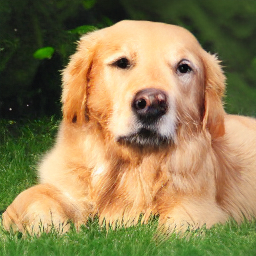, 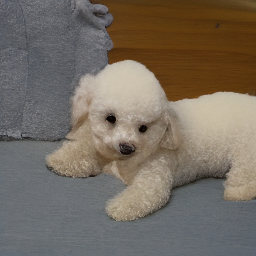, 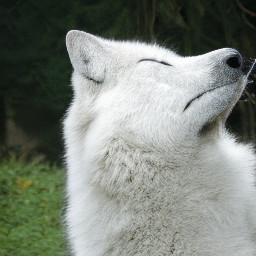, 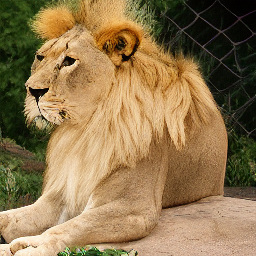, 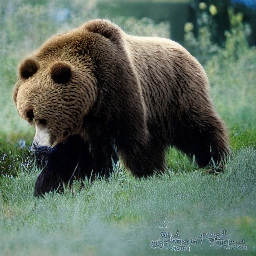, 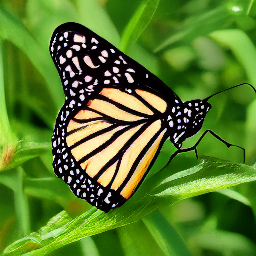, 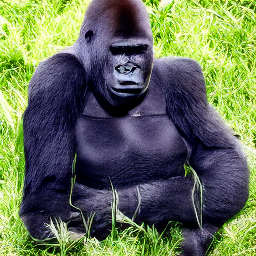, 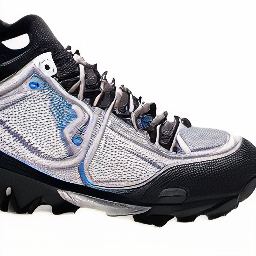, 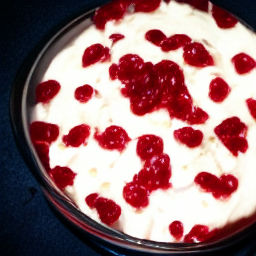, 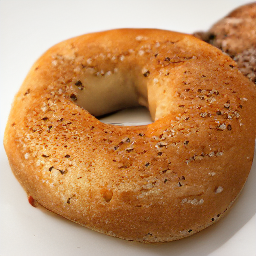, 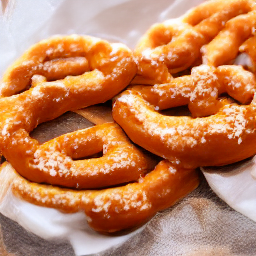, 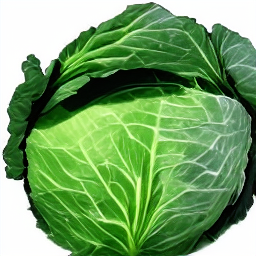, 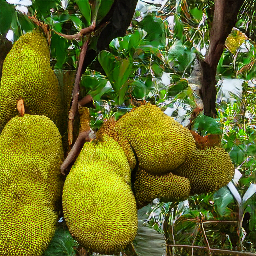, 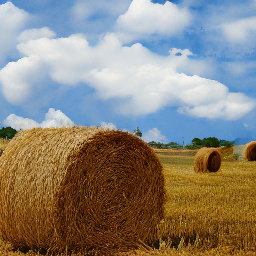, 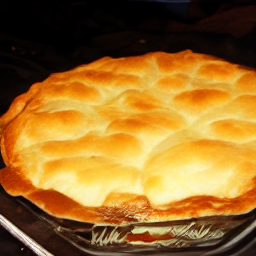, 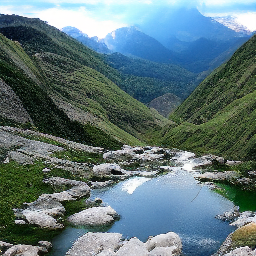, 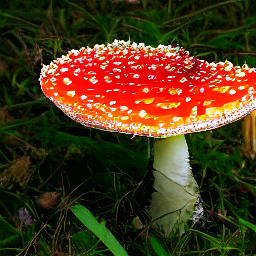
}
\newcounter{imgcounter}

\begin{figure}[htbp]
    \begin{adjustbox}{width=1.1\textwidth,center}  
        \begin{minipage}{\textwidth}
            \centering
            \setcounter{imgcounter}{0}
            \foreach \img in \imagelistbaseline{%
                \stepcounter{imgcounter}%
                \ifnum\value{imgcounter}>30\relax\else%
                    \begin{subfigure}[b]{0.19\textwidth}
                        \centering
                        \includegraphics[width=\textwidth]{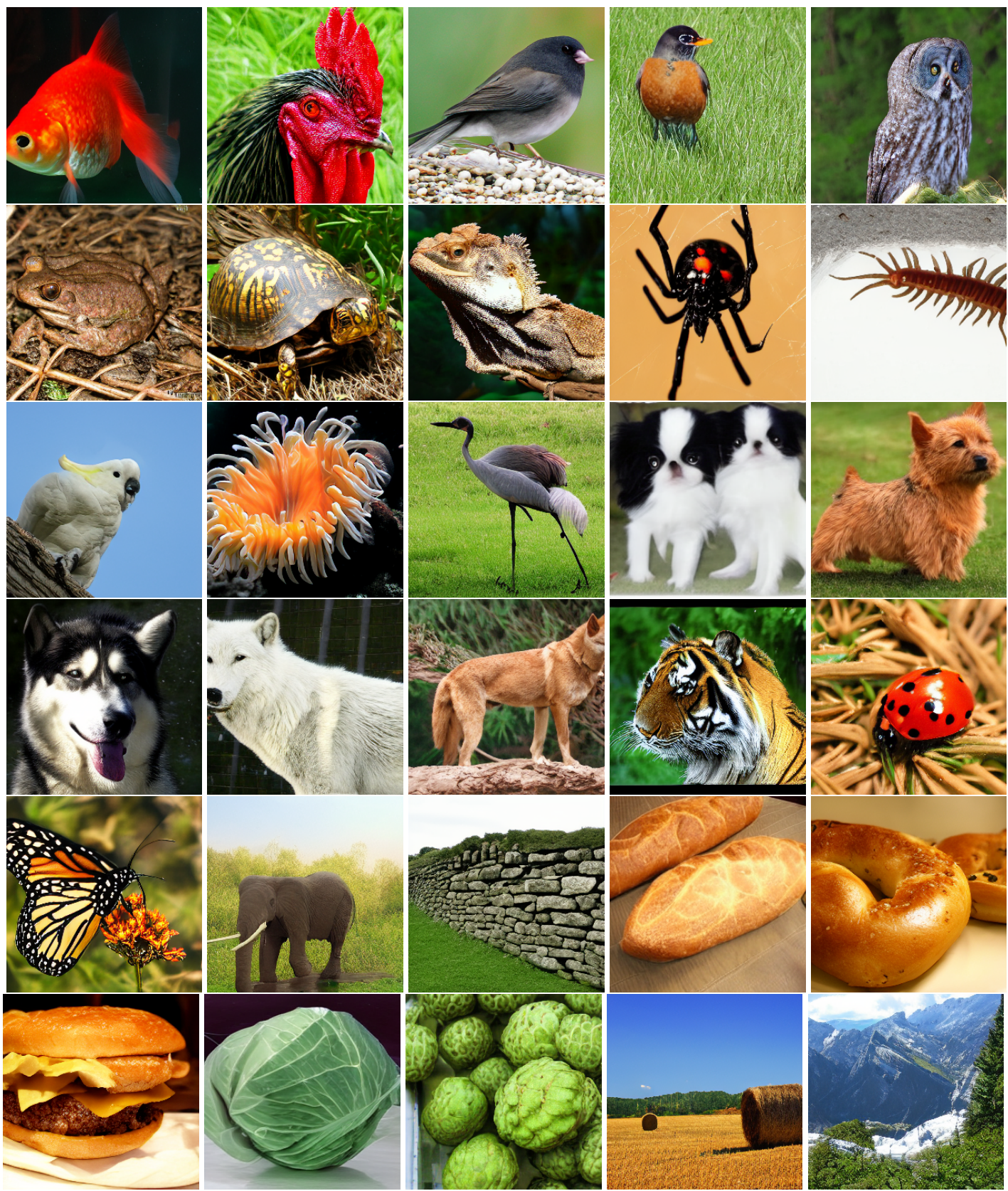}
                    \end{subfigure}%
                    \hspace{1mm}%
                \fi
                }
        \end{minipage}
    \end{adjustbox}
    \caption{ImageNet Samples for baseline model trained over $1,000$ latent and then sampling done for $\text{S}=64$ sampling steps.}
    \label{fig:allfigures}
\end{figure}

\newpage

\begin{figure}[htbp]
    \begin{adjustbox}{width=1.1\textwidth,center}  
        \begin{minipage}{\textwidth}
            \centering
            \setcounter{imgcounter}{0}
            \foreach \img in \imagelistsingle{%
                \stepcounter{imgcounter}%
                \ifnum\value{imgcounter}>30\relax\else%
                    \begin{subfigure}[b]{0.19\textwidth}
                        \centering
                        \includegraphics[width=\textwidth]{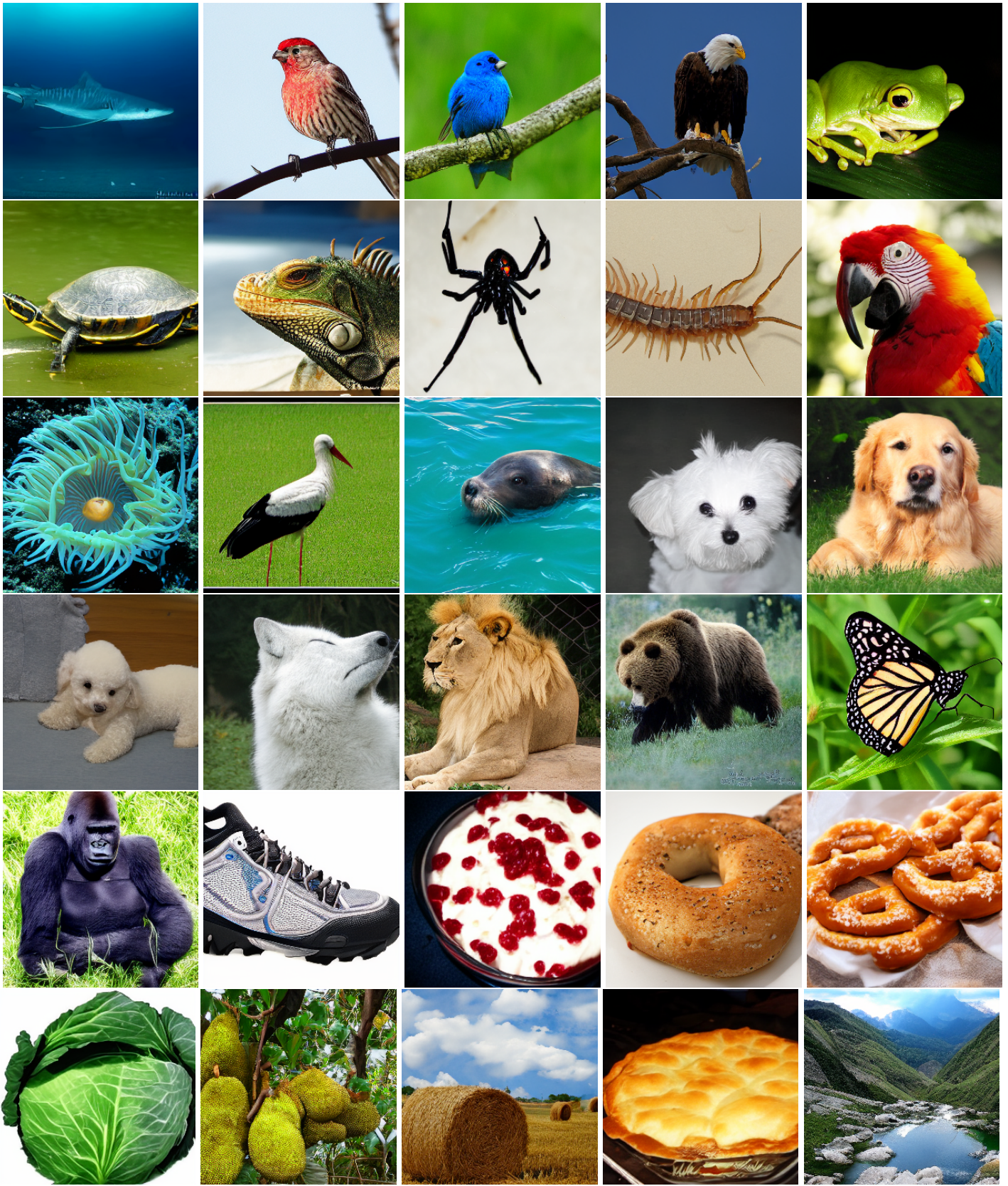}
                    \end{subfigure}%
                    \hspace{1mm}%
                \fi
                }
        \end{minipage}
    \end{adjustbox}
    \caption{ImageNet Samples for fewer latent-state model, i.e. a model trained only on  $\text{S}=64$ latent states and then sampling done on these $64$ latent-states as well.}
    \label{fig:allfigures}
\end{figure}

\begin{figure}[htbp]
    \begin{adjustbox}{width=1.1\textwidth,center}  
        \begin{minipage}{\textwidth}
            \centering
            \setcounter{imgcounter}{0}
            \foreach \img in \imagelistdisen{%
                \stepcounter{imgcounter}%
                \ifnum\value{imgcounter}>30\relax\else%
                    \begin{subfigure}[b]{0.19\textwidth}
                        \centering
                        \includegraphics[width=\textwidth]{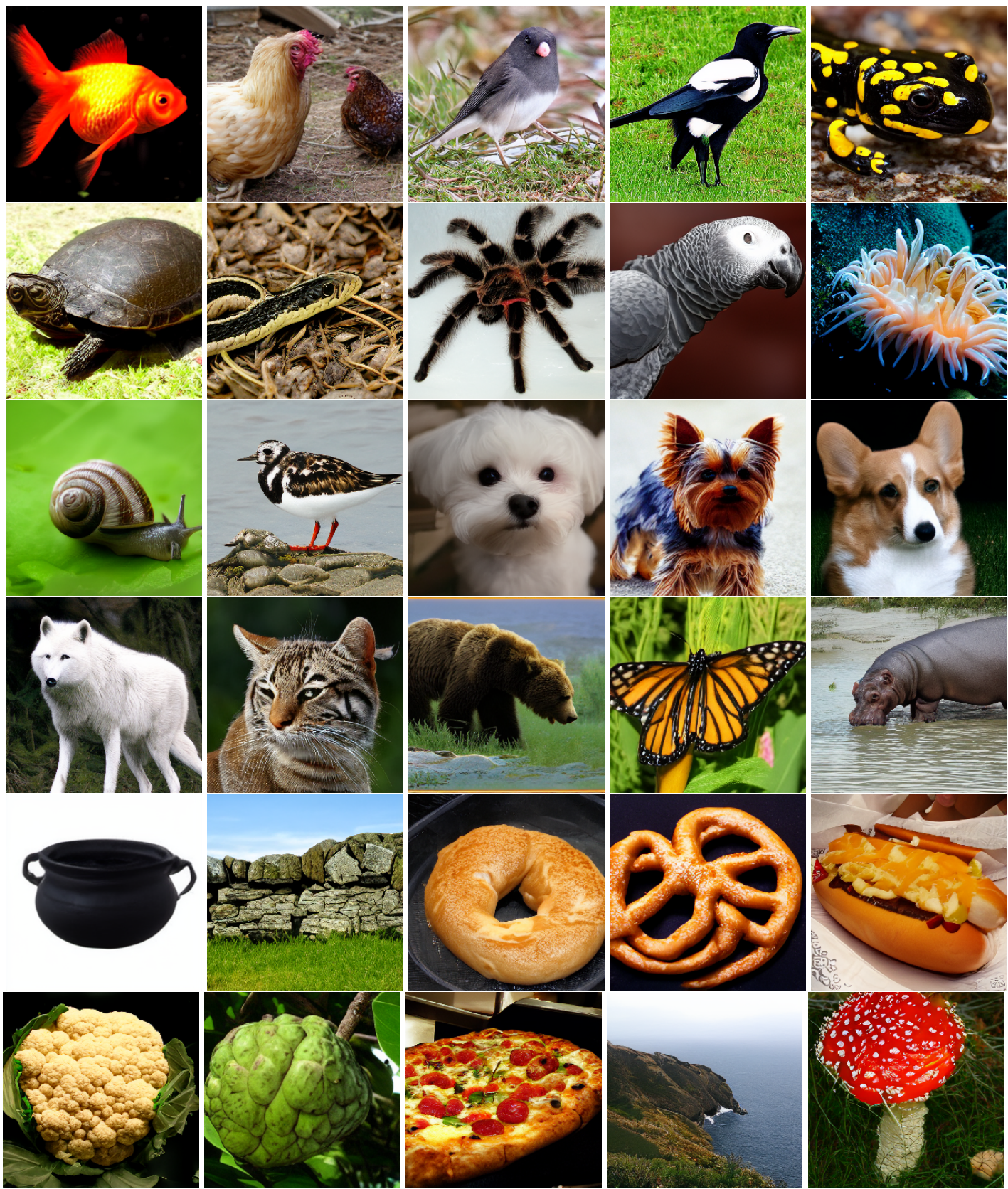}
                    \end{subfigure}%
                    \hspace{1mm}%
                \fi
                }
        \end{minipage}
    \end{adjustbox}
    \caption{ImageNet Samples for disentangled model for $\text{S}=64$.}
    \label{fig:allfigures}
\end{figure}

\clearpage
\subsection{Samples from different models}
We provide samples from various models discussed in the paper in a 3x3 tiled image:
\begin{itemize}
    \item The three columns correspond to different number of DDIM sampling steps.\\
\item The first row corresponds to different disentangled models obtained by combining $\text{S} $ different single latent-state models using algorithm \ref{algo:limit_training}. Therefore, R1C1 corresponds to model with 865x8 M parameters, R1C2 corresponds to model with 865x16 M parameters and R1C3 corresponds to model with 865x32M parameters.  Note that all combined resulting disentangled models are trained only for a single day (in parallel).\\
 \item The second row corresponds to the baseline model (with 865M parameters) i.e. standard text-to-image diffusion model trained over $1000$ latent states using algorithm \ref{algo:ddpm_training}. Here exactly same weights are used to sample for different sampling steps. This model is trained for 5 days.\\
\item The third row corresponds to the fewer latent-state models i.e. trained only on $\text{S}$ corresponding latent states using algorithm \ref{algo:rev_training}. Therefore R3C1 corresponds to model (with 865M parameters) trained over $\text{S}=8$ latent-states trained from scratch, R3C2 corresponds to a model (with 865M parameters) trained over $\text{S}=16$ latent states trained from scratch and R3C3 corresponds to a model (with 865M parameters) trained over $\text{S}=32$ latent states trained from scratch. All the three models have been trained for 5 days.
\end{itemize}

\begin{figure}[htbp]
\vspace{-0.5cm}
    \centering
    \includegraphics[ scale=1.0]{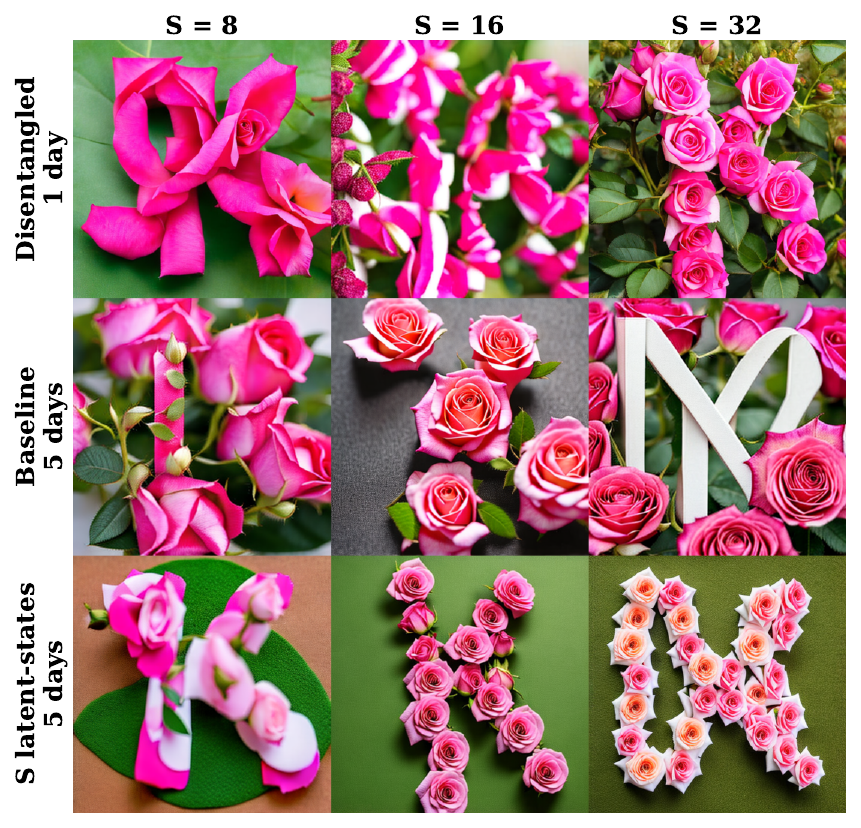}
    \caption{Prompt: Close up of letter K made out of pink roses.}

\end{figure}

\begin{figure}[htbp]
    \centering
    \includegraphics[scale=1.0]{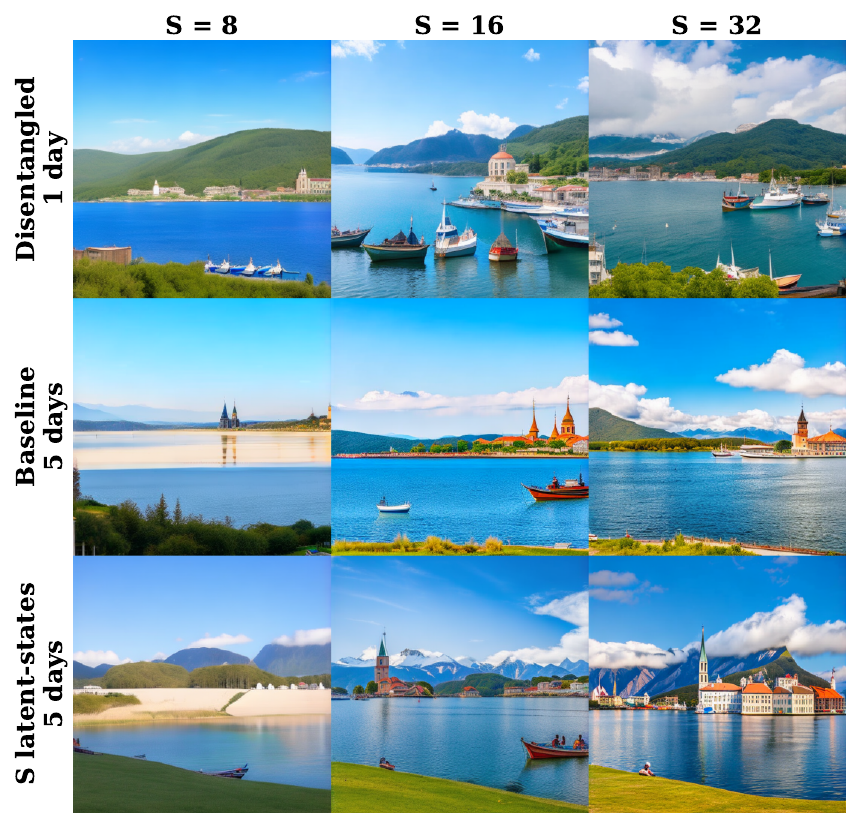}
    \caption{Prompt: View of body of water with boats and mountains in the background.}

\end{figure}

\begin{figure}[htbp]
    \centering
    \includegraphics[scale=1.0]{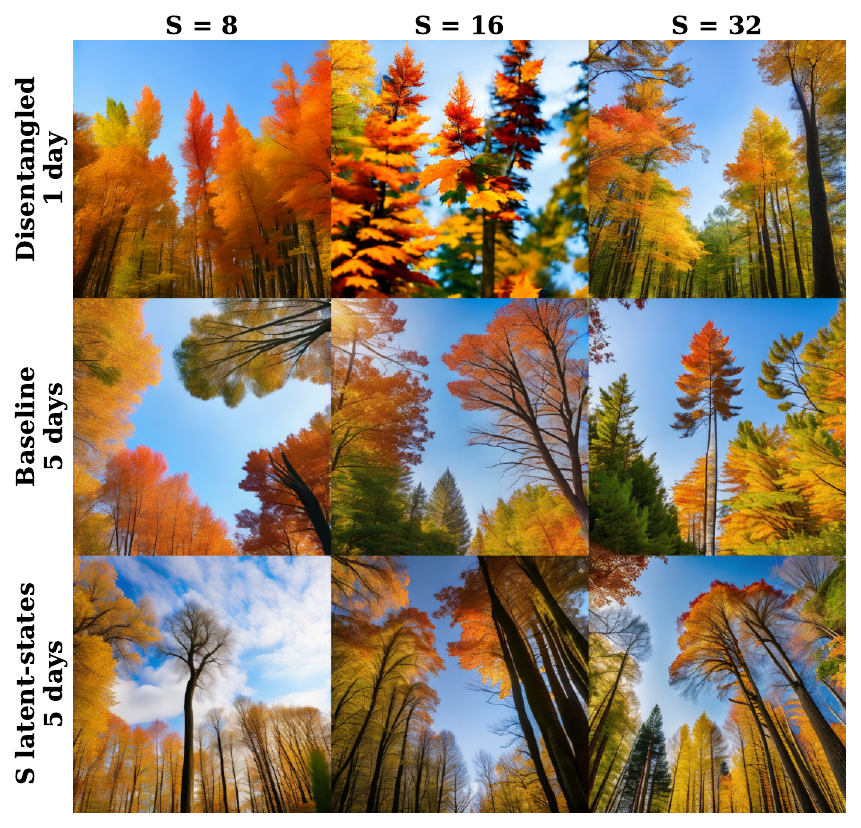}
    \caption{Prompt: Forest of tall trees with bright orange and yellow leaves against a clear blue sky.}

\end{figure}

\begin{figure}[htbp]
    \centering
    \includegraphics[scale=1.0]{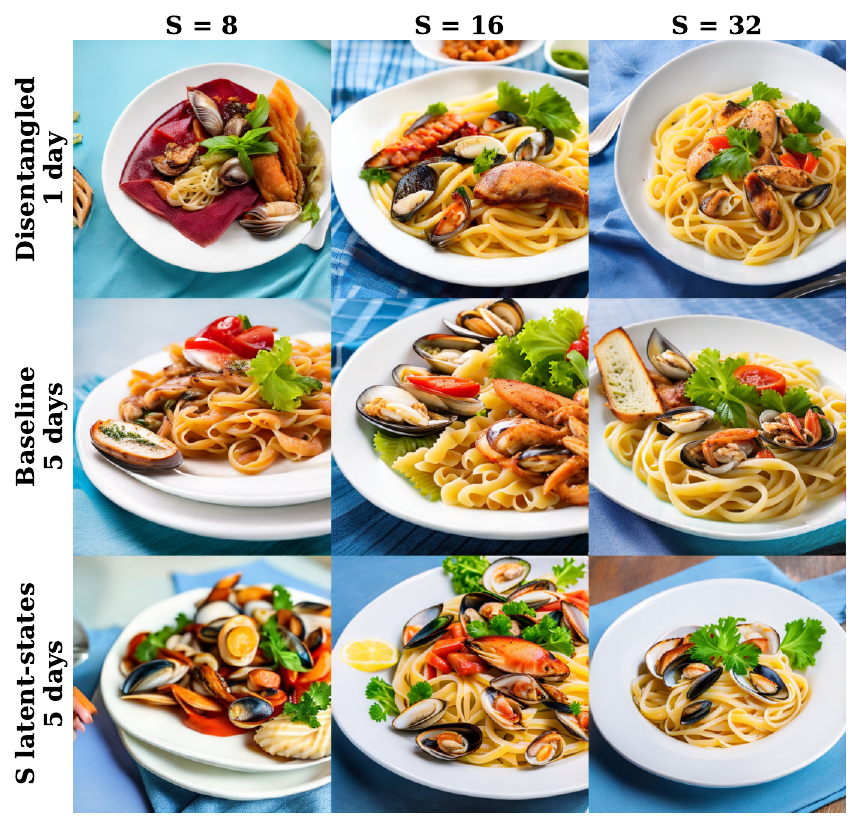}
    \caption{Prompt: Dish of pasta and seafood served in white plate on a blue tablecloth.}

\end{figure}

\begin{figure}[htbp]
    \centering
    \includegraphics[scale=1.0]{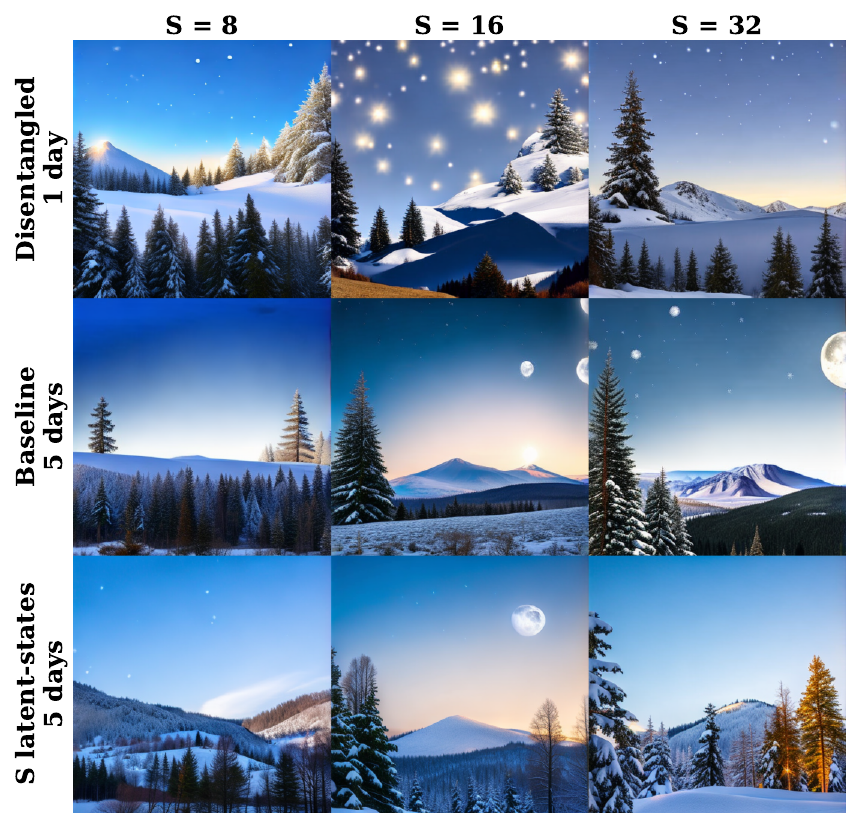}
    \caption{Prompt: Snow covered mountain with tall trees in the foreground and shining stars in the background.}

\end{figure}

\begin{figure}[htbp]
    \centering
    \includegraphics[scale=1.0]{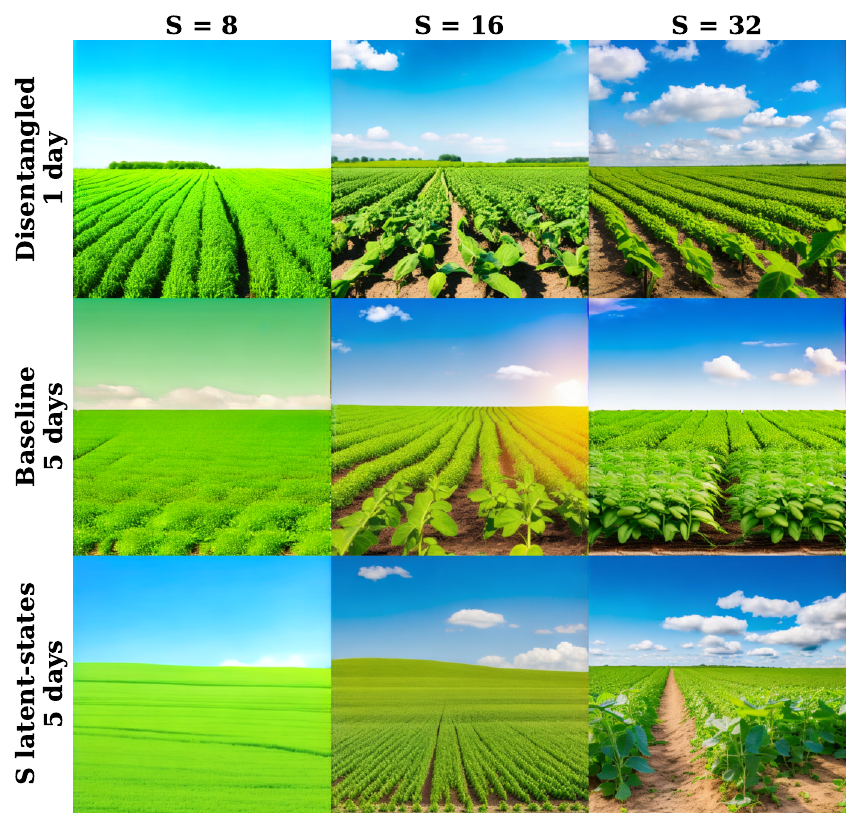}
    \caption{Prompt: Field of green soybeans growing in rows under clear blue sky with white fluffy clouds.}

\end{figure}

\begin{figure}[htbp]
    \centering
    \includegraphics[scale=1.0]{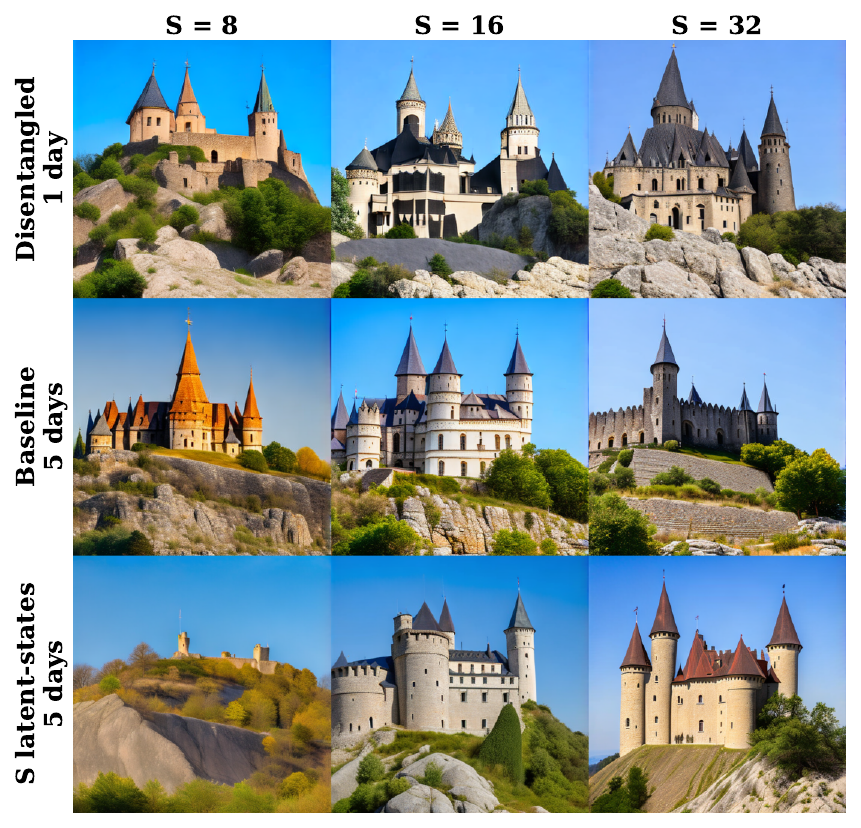}
    \caption{Prompt: Large dark castle sitting on top of a rocky hill. The castle has many spires and turrets on top and is made of stone.}

\end{figure}

\begin{figure}[htbp]
    \centering
    \includegraphics[scale=1.0]{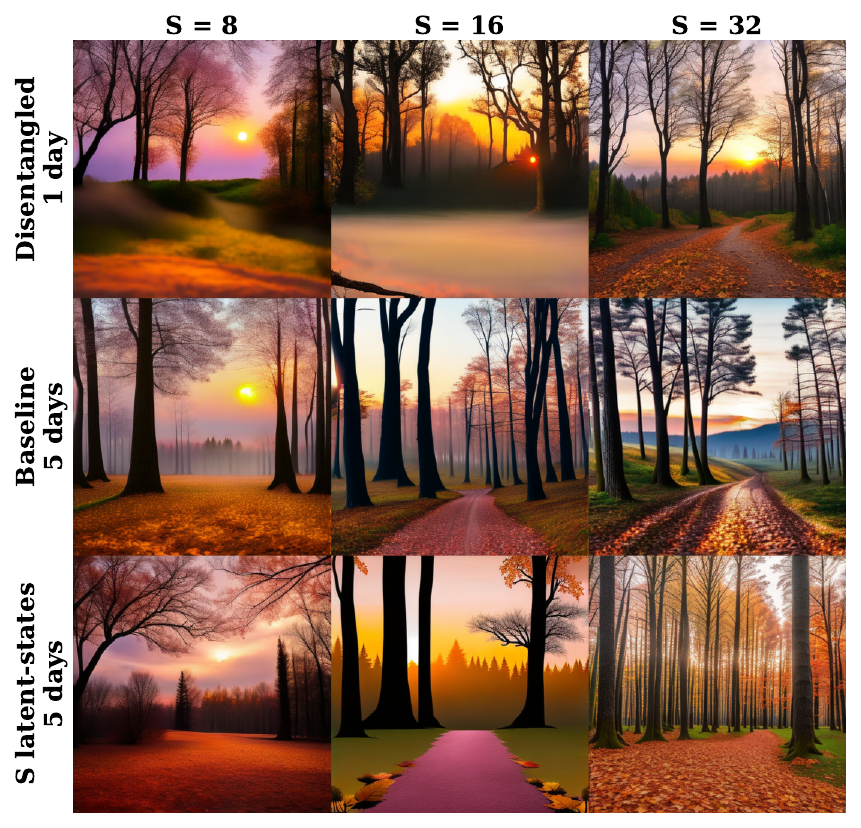}
    \caption{Prompt: Depiction of a forest at sunset, autumn colors, clouds in sky, fallen leaves.}

\end{figure}

\begin{figure}[htbp]
    \centering
    \includegraphics[scale=1.0]{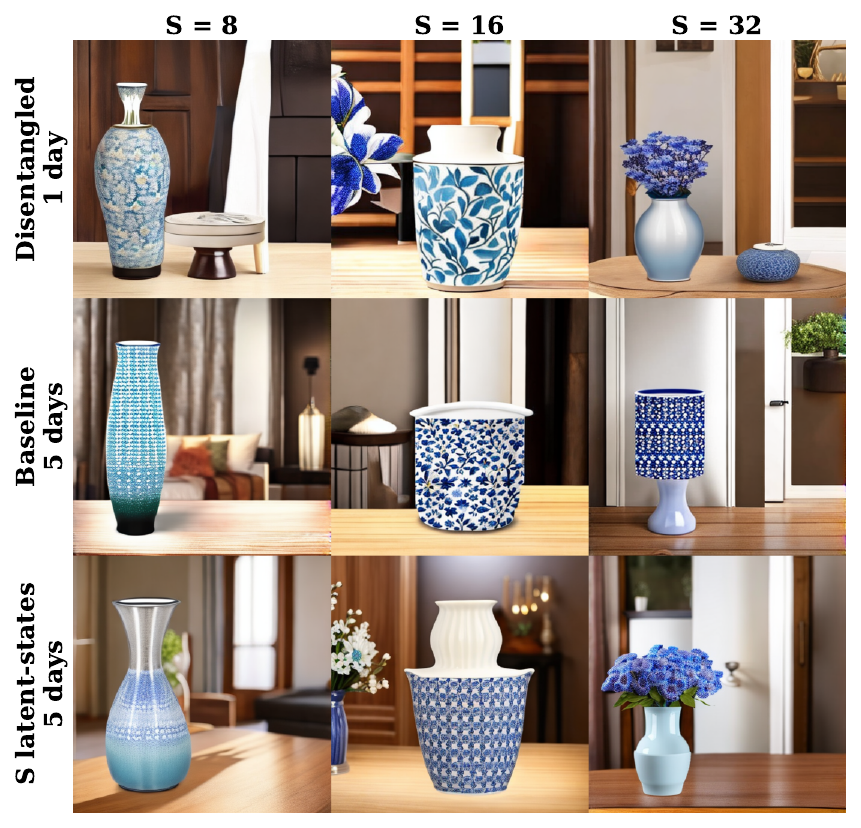}
    \caption{Prompt: Vase with blue and white floral pattern on it and placed on a wooden surface.}

\end{figure}

\begin{figure}[htbp]
    \centering
    \includegraphics[scale=1.0]{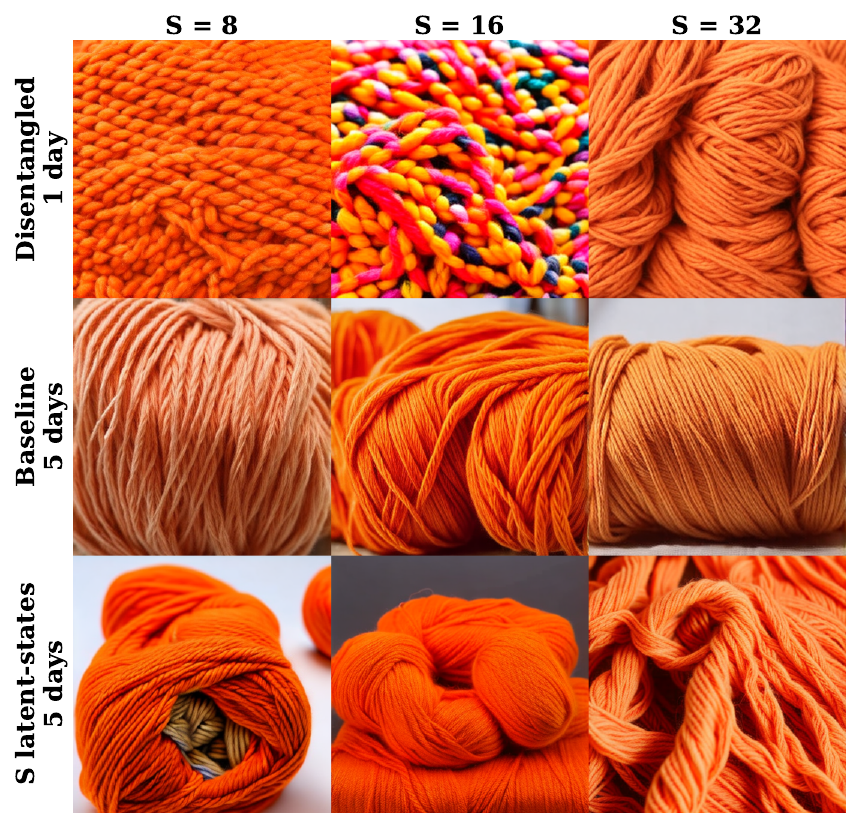}
    \caption{Prompt: Bright orange yarn knitted into a loop made of soft, wool like fibers and twisted together to form a dense bulky texture.}

\end{figure}

\begin{figure}[htbp]
    \centering
    \includegraphics[scale=1.0]{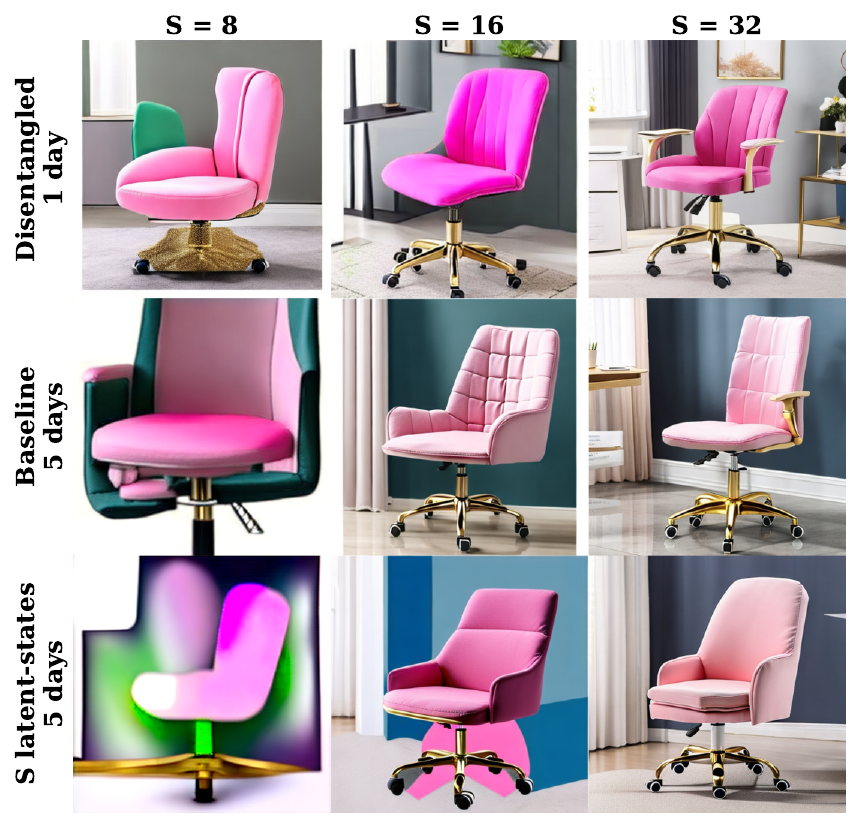}
    \caption{Prompt: Pink office chair with golden base and white casters. }

\end{figure}

\clearpage
\subsection{Latent-states}

The latent-states used in the paper are linearly spaced between $0$ and $999$, and are as follows:
\begin{align*}
S_4 &= \{249, 499, 749, 999\} \\
S_8 &= \{124, 249, 374, 499, 624, 749, 874, 999\} \\
S_{16} &= \{62, 124, 186, 249, 311, 374, 436, 499, 561, 624, 686, 749, 811, 874, 936, 999\} \\
S_{32} &= \{ 31,62,93,124,155,186,217,249,280,311,342,374,405,436,467,499, \\
& \phantom{111,}530,561,592,624,655,686,717,749,780,811,842,874,905,936,967,999 \} \\
S_{64} &= \{  16, 31, 46, 62, 77, 93, 108, 124, 139, 155, 170, 186, 201, 217, 233, 249, \\
 & \phantom{111,} 280, 295, 311, 326, 342, 358, 374, 389, 405, 420,
             436, 451, 467, 483, 499, \\
             & \phantom{111,} 530, 545, 561, 576, 592, 608, 624, 639, 655, 670, 686, 701, 717, 733, 749,\\
             & \phantom{111,} 764, 780, 795, 811, 826,
             842, 858, 874, 889, 905, 920, 936, 951, 967, 983, 999\} \hspace{2cm}
\end{align*}